\documentclass{article}

\usepackage{microtype}
\usepackage{graphicx}
\usepackage{subfigure}
\usepackage{booktabs} 
\usepackage{amsmath}
\usepackage{svg} 

\usepackage{hyperref}

\usepackage[accepted]{icml2022}

\icmltitlerunning{SPECTRE: Spectral Conditioning Helps to Overcome the Expressivity Limits of One-shot Graph Generators}

\usepackage{bm}
\usepackage{siunitx}
\usepackage{graphicx}
\usepackage{amsfonts}
\usepackage{xspace}
\usepackage{balance}

\usepackage{color, colortbl}
\definecolor{Gray}{gray}{0.9}
\definecolor{LightGray}{gray}{0.97}
\newcolumntype{g}{>{\columncolor{LightGray}}S}

\newcommand{\name}{SPECTRE\xspace}

\usepackage{mathbbol}
\usepackage{amssymb}             
\DeclareSymbolFontAlphabet{\amsmathbb}{AMSb}%

\newcommand{\diag}{\text{diag}}

\newcommand{\bb}[1]{\amsmathbb{#1}}
\newcommand{\bbR}{\bb{R}}
\newcommand{\bbLambda}{\mathbb{\Lambda}}

\newcommand{\GO}[1]{\bb{O}(#1)}
\newcommand{\SO}[1]{\bb{SO}(#1)}
\newcommand{\V}[2]{\bb{V}_{#1}(#2)}

\newcommand{\bA}{\bm{A}}
\newcommand{\bU}{\bm{U}}
\newcommand{\bR}{\bm{R}}
\newcommand{\bB}{\bm{B}}
\newcommand{\bQ}{\bm{Q}}

\newcommand{\bLambda}{\bm{\Lambda}}
\newcommand{\blambda}{\bm{\lambda}}

\newcommand{\bL}{\bm{L}}
\newcommand{\bI}{\bm{I}}
\newcommand{\bD}{\bm{D}}
\newcommand{\bM}{\bm{M}}

\newcommand{\bz}{\bm{z}}
\newcommand{\bw}{\bm{w}}
\newcommand{\bu}{\bm{u}}

\newcommand{\bx}{\bm{x}}
\newcommand{\bX}{\bm{X}}
\newcommand{\bY}{\bm{Y}}
\newcommand{\bS}{\bm{S}}

\renewcommand{\cal}[1]{\mathcal{#1}}
\newcommand{\cV}{\cal{V}}
\newcommand{\cE}{\cal{E}}

\begin{document}

\twocolumn[
\icmltitle{SPECTRE: Spectral Conditioning Helps to Overcome the Expressivity Limits of One-shot Graph Generators}



\icmlsetsymbol{equal}{*}

\begin{icmlauthorlist}
\icmlauthor{Karolis Martinkus}{to}
\icmlauthor{Andreas Loukas}{equal,goo}
\icmlauthor{Nathana\"{e}l Perraudin}{equal,ed}
\icmlauthor{Roger Wattenhofer}{to}
\end{icmlauthorlist}

\icmlaffiliation{to}{ETH Zurich}
\icmlaffiliation{goo}{EPFL and Prescient Design, Genentech}
\icmlaffiliation{ed}{Swiss Data Science Center}

\icmlcorrespondingauthor{Karolis Martinkus}{martinkus@ethz.com}

\icmlkeywords{Machine Learning, ICML}

\vskip 0.3in
]



\printAffiliationsAndNotice{\icmlEqualContribution} 

\begin{abstract}
We approach the graph generation problem from a spectral perspective by first generating the dominant parts of the graph Laplacian spectrum and then building a graph matching these eigenvalues and eigenvectors. Spectral conditioning allows for direct modeling of the global and local graph structure and helps to overcome the expressivity and mode collapse issues of one-shot graph generators.
Our novel GAN, called SPECTRE, enables the one-shot generation of much larger graphs than previously possible with one-shot models. SPECTRE outperforms state-of-the-art deep autoregressive generators in terms of modeling fidelity, while also avoiding expensive sequential generation and dependence on node ordering. A case in point, in sizable synthetic and real-world graphs SPECTRE achieves a 4-to-170 fold improvement over the best competitor that does not overfit and is 23-to-30 times faster than autoregressive generators.
\end{abstract}

\section{Introduction}

The ability to generate new samples from a distribution is a central problem in machine learning. Most of the work has focused on data with a regular structure, such as images and audio~\cite{brock2018large,oord2016wavenet}. For such data, Generative Adversarial Networks (GANs)~\cite{goodfellow2014generative,arjovsky2017wasserstein} have emerged as a powerful paradigm, managing to balance generation novelty and fidelity in a manner previously thought impossible. 

The present work considers the use of GANs for graph data. The generation of novel graphs is relevant for numerous applications in molecule~\cite{jin2018junction, de2018molgan, liu2018constrained}, protein~\cite{huang2016coming}, network, and circuit design~\cite{mirhoseini2021graph}. Yet, despite recent efforts, the problem remains largely unresolved: current state-of-the-art approaches are constrained to small graphs and often fail to strike a beneficial trade-off between capturing the essential properties of the training distribution and exhibiting high novelty. 
We argue that the one-shot generators used in current GAN models face expressivity issues that hinder them from capturing the global graph properties. We refer to a model as one-shot if it generates all edges between nodes at once. In contrast, auto-regressive models build a graph by progressively adding new nodes and edges. 

\begin{figure}[t!]
\includegraphics[width=\linewidth]{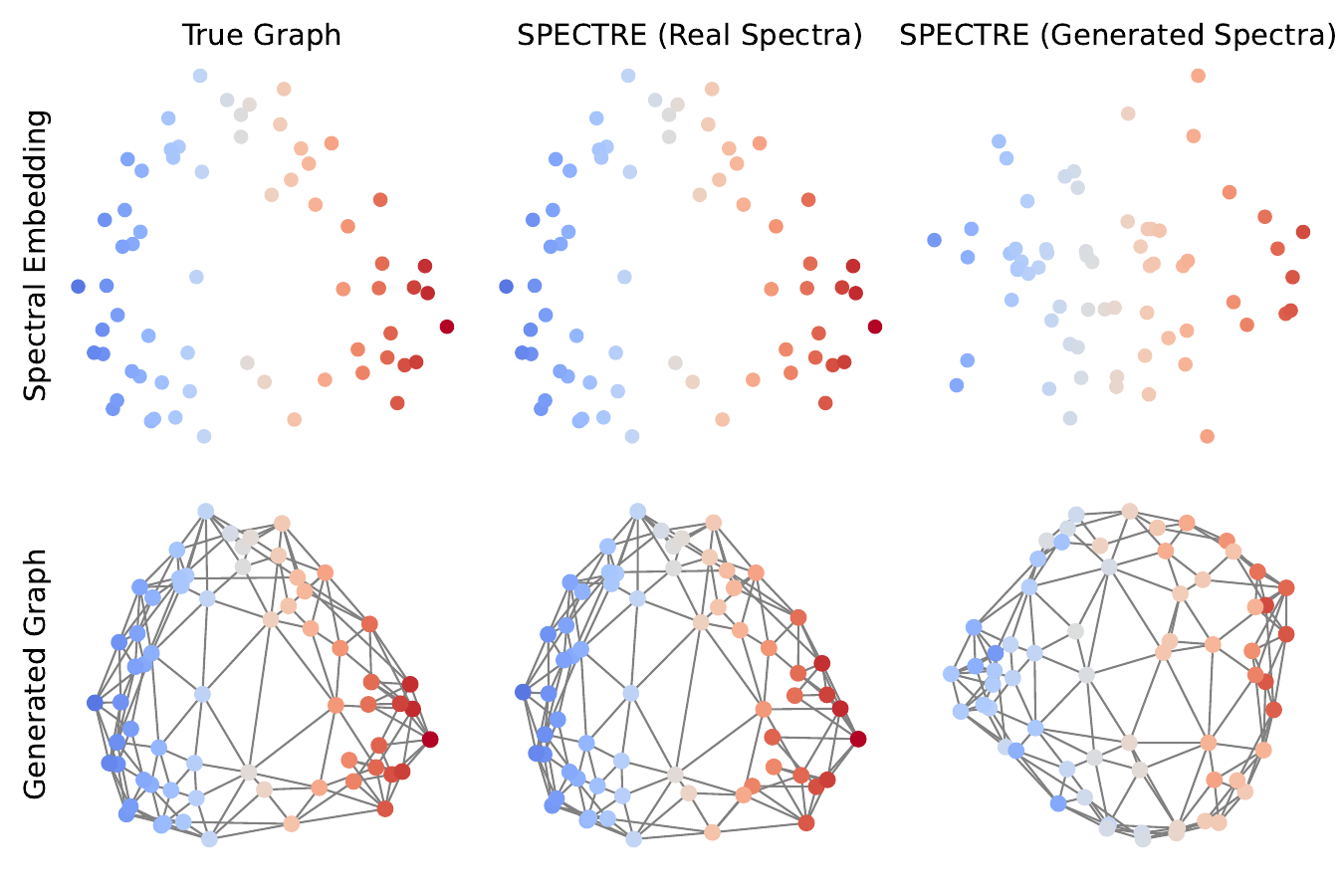}
\caption{Generating the spectrum first allows \name to control the global graph structure prior to the local connectivity. 
\name is also able to correct for imperfect generated spectra.
\textbf{Top}: Conditioning spectral embedding. \textbf{Bottom}: The generated graph, plotted using its final spectral embedding.
}
\label{fig:bait}
\end{figure}

One-shot generators are typically confronted with the demanding task of controlling global graph structure emergent from a large number of local node interactions. 
As a thought experiment, consider the toy problem of generating a tree by sampling node embeddings and then connecting them using some similarity kernel~\cite{krawczuk2020gg,serviansky2020set2graph,vignac2021top}. Even if the extended neighborhood of each node is locally tree-like, the overall graph will not be valid unless \textit{all} nodes are positioned appropriately w.r.t. each other. Misplacing even a few nodes can completely alter the global properties by introducing cycles and rendering the graph disconnected. Sadly, the aforementioned expressivity issue becomes increasingly pronounced as the size of the graph grows and can manifest in terms of non-convergent GAN training. 

This work puts forth \name---an equivariant generator that aims to overcome the expressivity issues of one-shot approaches. \name decomposes the graph generation problem into two parts that are learned jointly: (\textit{i}) modeling the dominant components of the graph spectrum and (\textit{ii}) generating a graph conditioned on a set of eigenvalues and eigenvectors.  
Modeling the distribution of the top-$k$ eigenvalues and eigenvectors is a simpler problem than one-shot graph generation.
Crucially, a direct inspection of the graph spectrum conveys many pertinent global graph properties (e.g., connectivity, cluster structure, diameter) and can be utilized to construct embeddings that approximate the geodesic distance between nodes~\cite{belkin2003laplacian,coifman2005geometric,mohar1997some}. As shown in Figure~\ref{fig:bait}, the (normalized) graph Laplacian eigenvectors (associated with low eigenvalues) capture well the coarse graph  structure, making them ideal to model non-local dependencies.  Thus, by generating the spectrum first, \name can control the global properties of the generated graphs. The learned eigenvectors and eigenvalues are then used to initialize the node embeddings of a second generator (inspired by GG-GAN~\cite{krawczuk2020gg}) that acts as a local refinement procedure.
Both steps are permutation equivariant, differentiable, and are optimized jointly in an end-to-end fashion. 

Our experiments with synthetic and real-world graphs provide evidence that spectral conditioning helps to overcome the limitations of one-shot generators, managing to faithfully capture the graph statistics even for graphs with hundreds of nodes. Interestingly, \name can outperform state-of-the-art by a non-negligible margin, striking a compelling trade-off between the ability to generate graphs not in the training distribution (novelty) and modeling fidelity. We also find that conditioning \name on real spectra can yield further improvement without sacrificing novelty, especially when $k$ is large, indicating that additional gains may be attainable by using a better spectrum generator.

\section{Related Work}

Graph generation entails building statistical models and fitting them to distributions of graphs. In this light, the problem is intimately linked with random graph models~\cite{erdos1960evolution,holland1983stochastic,eldridge2016graphons}  studied extensively in discrete mathematics, statistical physics, and network science. These models represent useful abstractions but are generally too simplistic to fit the real graph distributions that we care about. We refer the interested reader to the surveys~\cite{chakrabarti2006graph,goldenberg2010survey} for a more in-depth exposition.

Deep learning approaches, on the other hand, forego the simplicity and tractability of most random graph models to achieve greater data fidelity. Two main types of deep learning approaches have been proposed: 

\textbf{Autoregressive.} {Autoregressive models learn to build a graph by iteratively adding new nodes and predicting their edges}~\cite{you2018graphrnn,liao2019efficient,dai2020scalable}. By breaking the problem into smaller manageable parts, these methods are generally apt at capturing complex statistics of the data distribution. Unfortunately, training autoregressive generators hinges on imposing an order on the nodes, which renders the generator permutation sensitive and can lead to low novelty (due to memorizing the training set and generating different permutations of the same adjacency matrix). In addition, autoregressive generators are generally non fully parallelizable and thus take longer to generate larger graphs.
 
\textbf{One-shot.} {One-shot models aim to learn how to generate all edges between nodes at once, bringing the promise of parallelization}. Within one-shot methods we can distinguish those that build on the variational autoencoder (VAE) and generative adversarial network (GAN) paradigms: 
Graph VAEs~\cite{kipf2016variational,simonovsky2018graphvae,mitton2021graph} are generally easy to train and can work well for small graphs. A key challenge with VAEs that have a vector-based latent state is that to decouple the learning process from the permutation chosen when representing the graph as an adjacency matrix, one needs to solve a graph matching problem to align the VAE's input and output. The high computational complexity of graph matching 
necessitates the use of rough heuristics when training models with more than a handful of nodes and can significantly deteriorate performance.

The GAN framework provides an elegant alternative to VAEs~\citep{wang2018graphgan,de2018molgan,bojchevski2018netgan,yang2019conditional,krawczuk2020gg}. Adversarial training is believed to encourage larger sample diversity than maximizing likelihood, and moreover, by employing a permutation invariant discriminator one sidesteps the need for graph matching. 
Nevertheless, 
current equivariant one-shot GANs may exhibit convergence issues requiring involved fixes~\citep{yang2019conditional} and, even after recent improvements on their generator~\cite{krawczuk2020gg,vignac2021top}, they remain less apt at modeling complex graph statistics of larger graphs. Our work builds and improves upon this previous work: we find that GAN generators have difficulty in faithfully capturing graph statistics---a phenomenon that manifests more prominently as graphs become larger. We introduce the idea of spectrum conditioning to mitigate this challenge. 

Other generation strategies include learning a score function based on annealed Langevin dynamics~\citep{pmlr-v108-niu20a} and graph normalizing flows~\cite{liu2019graph}. 
We also mention two special instances of GANs: GraphGAN models the distribution of a node's neighborhood conditioned on the rest of the graph~\cite{wang2018graphgan}, whereas NETGAN assembles a graph from a collection of random walks generated by a GAN~\cite{bojchevski2018netgan}---these models were designed for (and evaluated on) graph representation learning tasks and not graph generation. VAEs have also been applied on such tasks \citep{shi2020effective}.

We note that the above discussion focuses on generally applicable methods and does not review specialized methods such as those specifically tailored to molecules~\cite{jin2018junction,you2018graph,liu2018constrained,bresson2019two,pmlr-v119-jin20a,garcia2021n,mahmood2021masked,liu2021graphebm}.

Finally, there exist optimization-based approaches for building co-spectral graphs~\cite{baldesi2018spectral,10.1145/3308558.3313631}. These works do not explicitly focus on modeling graph distributions.   

\section{Background}

In this contribution, we consider unweighted undirected graphs $G=(\cV, \cE)$ where $\cV$ is a set of $n$ nodes connected (or not) by a set of edges $\cE$. We index each node $ v_i \in \cV$ with $i=1,\ldots, n$ and define the adjacency matrix $\bA$ as $\bA_{i,j}=1$ if $v_i$ and $v_j$ are connected and $\bA_{i,j}=0$, otherwise. 

\subsection{Spectral Graph Theory}

Spectral graph theory studies the connections between the spectrum of the graph Laplacian and the general properties of the graph. 
The normalized graph Laplacian\footnote{Though, in the following, by graph Laplacian we refer to the normalized Laplacian, our ideas can be trivially extended to the combinatorial and random-walk Laplacian matrices.}  is defined as 
$
\bL= \bI - \bD^{-\frac{1}{2}} \bA \bD^{-\frac{1}{2}},
$
where $\bD = \diag(d_1, \cdots, d_n)$ is the diagonal degree matrix defined as $d_{i}=\sum_{j=1}^n{\bA_{i,j}}$. 
Being a symmetric positive semi-definite matrix, the graph Laplacian can always be diagonalized as $\bL = \bU \bLambda \bU^T,$ where the orthogonal matrix $\bU = [\bu_1, \cdots, \bu_n]$ and the diagonal matrix $\bLambda=\text{diag}(\lambda_1, \cdots, \lambda_n)$ contain the graph Laplacian eigenvectors and eigenvalues, respectively. We follow the convention of sorting eigenvalues in non-decreasing order as $0 = \lambda_1 \leq \lambda_2 \leq \cdots \leq \lambda_n \leq 2$.

Our approach is motivated by the well known fact that the first few eigenvalues $\blambda_k = (\lambda_1, \cdots, \lambda_k)$ and eigenvectors $\bU_k = [\bu_1, \cdots, \bu_k]$ of the graph Laplacian describe global properties of the graph structure such as its connectivity, clusterability, diameter, and node distance (see Appendix~\ref{app:spectrum}).  

We assume the graph to be connected. Therefore the first eigenvector (associated with $\lambda_1=0$) only contains information about the node degree (a local feature) and we start conditioning graph generation with the \emph{second} eigenvector.

\subsection{Orthogonal Matrices}
\label{subsec:orthogonal}

In the following, we describe some useful facts about the geometric, algebraic, and group structure of orthogonal matrices as eigenvectors of undirected graphs are such matrices. 

\textit{The special orthogonal group $\SO{n}$.} Being orthogonal matrices ($\bU \bU^\top = \bU^\top \bU = \bI$), eigenvectors belong to the orthogonal matrix group $\GO{n}$. The latter contains two connected components: one comprising of all of the matrices with a determinant of $-1$  and another comprising of all of the matrices with a determinant of $+1$ (rotation matrices), also known as the special orthogonal group $\SO{n}$. Graph eigenvectors are phase-independent, meaning that $-\bU$ corresponds to the same eigenbasis as $\bU$ and there always exists a $\SO{n}$ matrix that corresponds to a given graph's eigenbasis. We can thus generate all possible graph eigenspace while being restricted to $\SO{n}$, as we do in the following.

The Lie algebra of $\SO{n}$ is formed by skew-symmetric matrices ($\bS^T = -\bS$). The matrix exponential $\bU=\exp(\bS)$ can then be used as a surjective map from skew-symmetric matrices onto $\SO{n}$ \citep{shepard2015representation}. A simple counting argument reveals that orthogonal matrices can be defined using ${n(n-1)}/{2}$ parameters.

\textit{The Stiefel manifold $\V{k}{n}$.} In this work, we mostly focus on the first $k$ eigenvectors $\bU_{k} = [\bu_1, \cdots, \bu_k] \in \bbR^{n \times k}$, which form a Stiefel manifold $V_k(n)$ and abide to:
$
\bU_{k}^\top \bU_{k} = \bI_k,   
$
where $\bI_k$ is the $k \times k$ identity matrix. Note that $\bU_{k} \bU_{k}^\top = \bI_n$ iff $n=k$.
One point in the Stiefel manifold can be transformed into any other with a rotation, i.e. , using multiplication with orthogonal matrices $\bR_L \in \SO{n}$, $\bR_R \in \SO{k}$: 
\begin{equation} \label{eq:linear-transoform-rotation}
    \bU \in \V{k}{n} \Rightarrow \bR_L \bU \bR_R \in \V{k}{n}.
\end{equation}
We use this property to build the layers of the eigenvector generator (see Sec.~\ref{subsec:architecture_eigenvector}).
To generate a random initial point on the Stiefel manifold, one can create a random skew-symmetric matrix with $nk - {k(k+1)}/{2}$ non-zero parameters, compute its exponential and select the first $k$ columns.

\section{Spectrum and Graph Generation}

\begin{figure*}[ht]
\vskip 0.2in
\includegraphics[width=0.5\linewidth,trim=0 3mm 0 0, clip]{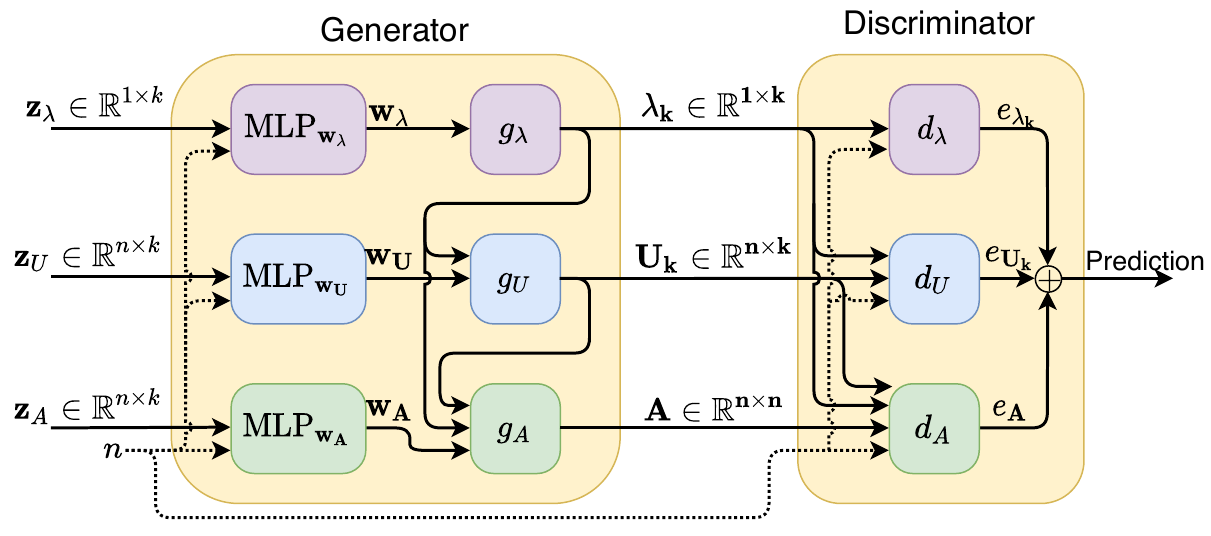}
\includegraphics[width=0.49\linewidth,trim=0 0 0 9mm, clip]{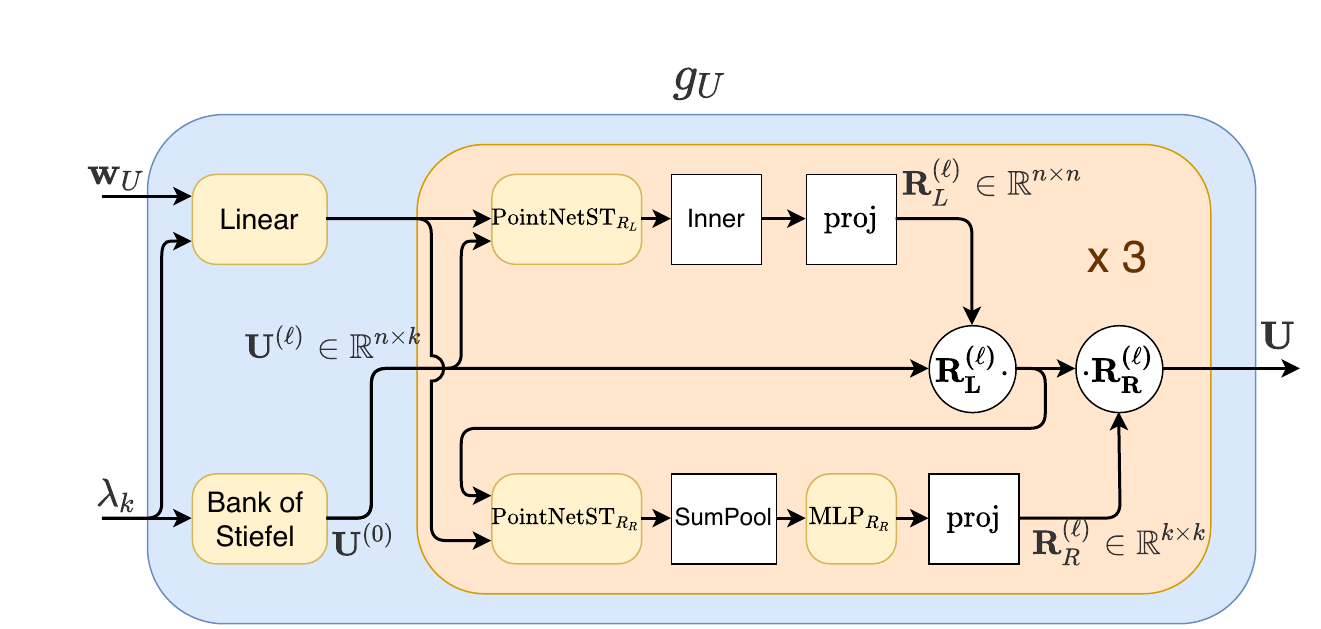}
\vskip -0.1in
\caption{\textbf{Left:} General architecture. Generation is performed sequentially with 3 GANs generating the eigenvalues (purple), the eigenvectors (blue) and finally the graph (green). Each generation step is conditioned on the previous generated variable. The input latent variable $\bw$ of each sub-generator is obtained using an MLP. \textbf{Right:} Eigenvector generator. The initial eigenvectors $\bU^{(0)}$ are selected from a learned Bank of Stiefel Manifolds. They are then transformed with  $3$ rotation layers (orange box) that each perform one left and one right rotation. }
\label{fig:architecture}
\end{figure*}

\name aims to overcome a key difficulty that one-shot graph generators face: controlling the global graph structure by manipulating local interactions.
To this end, \name generates graphs by first modeling the distribution of the top-$k$ eigenvalues and eigenvectors  $\blambda_k \in \bbLambda_k$  and $\bU_k \in \V{k}{n}$,
where $\bbLambda_k$ is the set of $k$ strictly positive non-decreasing eigenvalues of the graph Laplacian
$$
    \bbLambda_k := \{\blambda_k \in (0,2]^k \ \text{ s.t. } \  \lambda_1 \leq \cdots \leq \lambda_k\},
$$
whereas $\V{k}{n}$ is the Stiefel manifold. 

The graph is then generated conditioned on $(\bU_k, \blambda_k)$.
As explained in the background section, the dominant Laplacian spectra succinctly summarize the global structural properties of a graph and provide a rough embedding for the nodes. The latter can be used to bootstrap the graph generator module, simplifying its job.

We cover the architecture of SPECTRE (Figure \ref{fig:architecture}, left) by first presenting the conditional graph generator and then showing how the top-$k$ eigenvalue and eigenvector generation works. 
Each latent variable $\bz_\lambda \in \bbR^{1\times k}$, $\bz_U \in \bbR^{n\times k}$, $\bz_A\in \bbR^{n\times k}$ is obtained by sampling a uniform point $\bz_U$ from a hypersphere. Then, they are transformed using four-layer Multi-Layer Perceptrons ($\text{MLP}_{\bw_\lambda}$, $\text{MLP}_{\bw_U}$, $\text{MLP}_{\bw_A}$) to obtain $\bw_\lambda\in \bbR^{1\times k}$, $\bw_U\in \bbR^{n\times k}$, $\bw_A\in \bbR^{n\times k}$.
Further implementation details of the architecture can be found in Appendix~\ref{appx:architecture}.

\subsection{Conditional Graph Generation}

The \name graph generator $g_L$ aims to construct graphs with a given dominant spectra $(\blambda_k, \bU_k)$:
$$
     \bA = g_A(\blambda_k, \bU_k, \bw_A).
$$
First, we build $\bL^{(0)}\in \bbR^{n \times n}$, a rough approximation of the Laplacian matrix 
$$
    \bL^{(0)}  = \bU_k \, \diag(\blambda_k) \, \bU_k^\top \;.  
$$
Though $\bL^{(0)}$ does not look like a graph at this stage as it lacks the local connectivity information, it internally encodes the global graph structure. This matrix is passed in place of an adjacency matrix to a $l$-layer Provably Powerful Graph Network (PPGN) \citep{maron2019provably} using $\blambda_k$, $\bU_k$ and $\bw_A$ as node features. The PPGN then produces the final adjacency matrix.
We can interpret this as $g_A$ taking the initial approximation of the Laplacian matrix $\bL^{(0)}$ and progressively refining it: 
$$
   \bL^{(l)} = \text{PPGN}_l(\bL^{(l-1)}) \quad \text{for layer} \quad l = 1, \cdots, L-1  .
$$
The final layer then produces the adjacency matrix directly to avoid a complicated manual conversion from a Laplacian to an adjacency: 
$$
\bA =\sigma(\text{PPGN}_L(\bL^{(L-1)})),
$$
where $\sigma$ is the sigmoid activation function. Since we actually use high-dimensional representations between the PPGN layers, in this interpretation, this high-dimensional representation encodes $\bL^{(l)}$.
PPGN's expressive power matches that of a 3-WL test. The network is thus better suited to modeling and distinguishing graphs than typical Graph Neural Networks (GNNs) that typically are as discriminative as a 2-WL test~\citep{xu2018how,morris2019weisfeiler}. More recent GNNs are either less expressive than PPGN~\cite{zhang2021nested,sandfelder2021ego,papp2021dropgnn,bevilacqua2021equivariant}, or are computationally slower in the non-asymptotic regime~\cite{vignac2020building,NEURIPS2020_f81dee42}, or hinge on the computation of pre-defined features~\cite{bouritsas2020improving}. We stress that the list of mentioned references provides an incomplete sample of all known GNN architectures whose expressive power surpasses
 the 2-WL test.

We should note that the set $\bb{S}_k(n) \subset \bbLambda_k \times \V{k}{n}$ of \textit{valid graph Laplacian spectra}, i.e., those $(\blambda_k,\bU_k)$ that lead to a valid graph $G$, can be much smaller than those that are modeled by \name. Unfortunately, the problem of determining whether $(\blambda_k,\bU_k) \in \bb{S}_k(n)$ appears to be computationally hard when $k \ll n$, meaning that we cannot expect that the spectrum generator always returns a valid sample. We thus do not enforce exact conditioning on $(\blambda_k,\bU_k)$ but instead motivate the graph generator to generate graphs whose first $k$ eigenvalues and eigenvectors are close to those sampled during the first stages. 

The graph discriminator takes the adjacency matrix and corresponding spectral features as input:
$$
     e_{\bA} = d_A(\bA, \blambda_k, \bU_k, n) .
$$
Due to its conditioning, the discriminator helps to ensure that the generated graph and eigenvectors are consistent. To further encourage the discriminator to focus on the relation between the graph and the eigenvectors, we sometimes pass true (perturbed) eigenvalues and eigenvectors to the generator. 
The discriminator architecture is analogous to the generator, with an additional global pooling layer before the output. Note that having spectral features strengthens the discriminator. Consider, for instance, the task of determining whether a graph is connected---a task that a good discriminator needs to solve. Whereas in normal architectures the GNN depth needs to exceed the graph diameter to determine connectivity, when spectral features are available, connectivity can be checked locally. To guarantee global consistency, we then just need to ensure that the spectral features we condition on are well-posed.

Note that while in this paper we mainly focus on the generation of graph structure alone, we can also use the PPGN to directly produce node and edge features for the generated graph (see Appendix~\ref{appx:ppgn}).
 
\subsection{Conditional Eigenvector Generation}
\label{subsec:architecture_eigenvector}

The eigenvector generator (Figure~\ref{fig:architecture} right), aims to construct eigenvectors $\bU_k$ matching given eigenvalues $\blambda_k$:
$$
     \bU_k = g_U(\blambda_k, \bw_U),
$$
It operates by iteratively refining a starting eigenvector matrix. First, a starting Stiefel matrix $\bU_{k}^{(0)}$ is selected from a bank of learned matrices $\{\bB_0,...,\bB_m \}$ with $\bB_i \in \V{k}{n}$ for all $i = 1, \cdots, m$.
Using a bank helps to make the generation easier, as orthogonal matrices which correspond to valid graph Laplacian eigenvectors form a potentially small subset of all orthogonal matrices. However, it is possible to instead use only one fixed learnable rotation matrix as input, which would cause a slight decrease in generation quality.
The selection is done by generating a query matrix $\bQ = \text{MLP}(\blambda_k)$ of size ${n \times k}$ that is compared to matrices in the bank using the canonical Stiefel manifold metric~\citep{edelman1998geometry}:
$$
m(\bQ, \bB_i) := \text{tr}\big(\bQ^T\big(\bI-\frac{1}{2}\bB_{i}\bB_{i}^\top\big)\bB_{i} \big),
$$
which we normalize such that the distance from $\bB_{i}$ to itself is equal to one. Here $\text{tr}(X)$ computes the trace of the matrix. The starting matrix $\bU_{k}^{(0)}$ is then sampled using Gumbel-softmax \citep{jang2016categorical,maddison2016concrete}.

The generator proceeds to refine $\bU_{k}^{(0)}$ by repeatedly multiplying it with left and right orthogonal matrices:
$$
   \bU_{k}^{(\ell)} = \bR_{L}^{(\ell)} \, \bU_{k}^{(\ell-1)} \bR_{R}^{(\ell)}  \quad \text{for layer} \quad \ell = 1, \cdots, L.
$$
As described in \eqref{eq:linear-transoform-rotation} this transformation ensures that the matrix stays on the Stiefel manifold, meaning that it is a valid eigenvector matrix with orthogonal unit-length column vectors.
The left refinement matrix is constructed by processing inputs with a  PointNetST layer \citep{segol2019universal} and projecting the result onto $\SO{n}$ by constructing a skew-symmetric matrix and finally using the matrix exponential:
$$
\bR_L^{(\ell)} = \text{proj} \big( \text{outer} \big(\text{PointNetST}(\bU_{k}^{(\ell-1)}, \blambda_k, \bz_U)
\big)\big),
$$
where we define $\text{outer}(\bX) := \bX \bX^\top$ and
$ 
\text{proj}(\bX) := \exp( \text{tril}(\bX) - \text{tril}(\bX)^\top)
$, 
where $\text{tril}(X)$ gives the lower triangular of the matrix.
The right rotation matrix is constructed similarly by using a second PointNetST layer (no parameter-sharing) and mean pooling over the set of nodes:
$$
\bR_{R}^{(\ell)} = \text{proj} \big(
\text{MLP}_{\bR_R}(\frac{1}{n}\sum_{i=1}^n \text{PointNetST}(\bU_{k}^{(\ell-1)}, \blambda_k, \bz_U))
\big).
$$
The eigenvector discriminator takes the eigenvectors and the corresponding eigenvalues as input:
$$
     e_{\bU_k} = d_U(\bU_k, \blambda_k, n) .
$$
Since spectral node embeddings induce a clustering, for the discriminator we use an architecture based on PointNet \citep{qi2017pointnet}, which achieves good results on point cloud segmentation and classification. This architecture comprises of a right rotation, a point-wise transformation with an $\text{MLP}$, another right rotation, and a PointNetST layer followed by mean pooling and an $\text{MLP}$. The right rotations are constructed in the same way as done in the generator.

We use PointNetST layers as they are efficient and can approximate any equivariant set function \citep{segol2019universal}. Model weights are initialized such, that all of the rotation matrices are initially close to identity.

To avoid any sign ambiguity in the eigenvector representation we transform both the generated and the true eigenvectors such that the maximum absolute value for each eigenvector is positive.

\begin{figure*}[t!]
\centerline{
\includegraphics[width=0.99\linewidth]{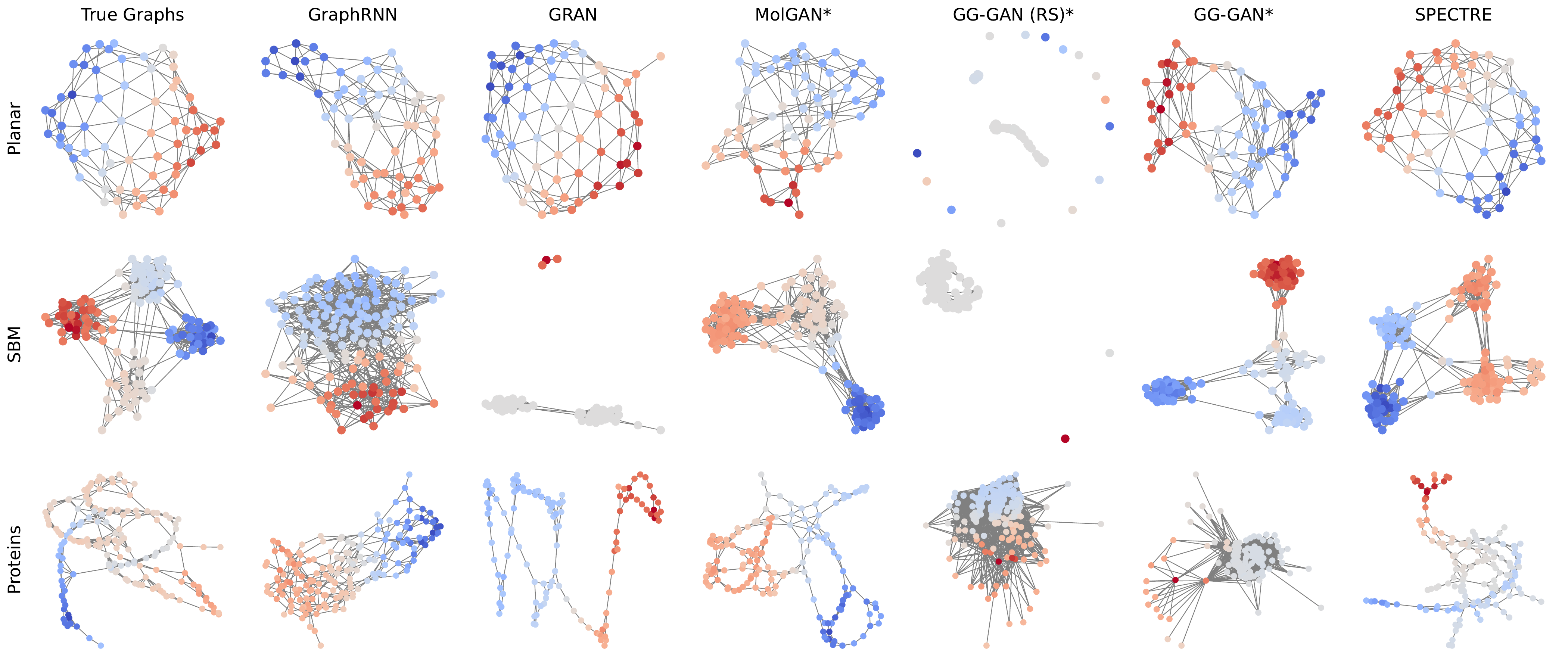}
}
\caption{A set of sample graphs produced by the models. Each row is conditioned on the same number of nodes.
}
\label{fig:graph_samples}
\end{figure*}

\subsection{Eigenvalue Generation}

The final piece of the puzzle entails generating 
$$
     \blambda_k = g_{\lambda}( \bw_{\lambda}).
$$
As eigenvalues are just an increasing sequence, the eigenvalue generator is a simple 4-layer 1D CNN with up-sampling \cite{donahue2018adversarial}.

Likewise, the discriminator:
$$
     e_{\blambda_k} = d_{\lambda}(\blambda_k, n) ,
$$
is a strided 4-layer 1D CNN with a linear final read-out layer. Both networks employ gated activation units $z = \tanh(W_{1}X) \cdot \sigma(W_{2}X)$ as used in WaveNet \citep{oord2016wavenet} and PixelCNN \citep{oord2016conditional}.

\subsection{Training}

To train our model we use the WGAN-LP loss ($\lambda_{\text{LP}} = 5$) \citep{petzka2018regularization}, ADAM optimizer ($\beta_1=0.5$, $\beta_2=0.9$) \citep{kingma2014adam}, and learning rate of $1e-4$ for both the generator and the discriminator. 

During training we utilize a form of teacher forcing, where initially each model is trained separately for 26k training steps, using true slightly perturbed spectral features from the training set for conditioning. We then gradually anneal the mixing temperature $\tau$, which defines how many real inputs each model receives, over the next 26k steps from $1.0$ to $0.8$ using a cosine schedule. This teacher forcing serves to ``teach'' the model how to construct graphs with given spectral features and correct small errors.
All of our models are trained for 150k steps in total. 
The code is publicly available\footnote{\url{https://github.com/KarolisMart/SPECTRE}}.

The selection of $k$ (the number of eigenvalues and eigenvectors) is done as follows. First we train only the graph generator $g_A$ conditioned on true spectra using different values of $k \in [2,4,8,16,32]$ for 26k steps. Then we select the lowest $k$ which resulted in the generation of good quality graph samples (low MMD measures).

\section{Experimental Evaluation}

Our experimental setup largely follows \citet{you2018graphrnn} and \citet{liao2019efficient} with some important extensions that are discussed below.

\paragraph{Datasets.} We consider five real and synthetic datasets of varying size and connectivity: Community-small (12-20 nodes)~\cite{you2018graphrnn}, QM9 ($9$ nodes)~\cite{ramakrishnan2014quantum}, Planar graphs ($64$ nodes), a Stochastic Block Model (20-40 nodes per community, 2-5 communities), Proteins (100-500 nodes)~\citep{dobson2003distinguishing}. All datasets are described in Appendix~\ref{app:datasets}. We split all datasets into test (20\%) and training (80\%) sets. We use 20\% of the training set for validation. We generate the same number of samples as there are in the test split of each dataset.

\begin{table*}[ht]
\centering
\resizebox{\linewidth}{!}{\begin{tabular}{l*{5}{S}g*{3}{S}gS}
\toprule
 & \multicolumn{11}{c}{Planar graphs}\\
\cmidrule(lr){2-12}
{Model} & {Deg.\,$\downarrow$} & {Clus.\,$\downarrow$} & {Orbit\, $\downarrow$} & {Spec.\,$\downarrow$} & {Wavelet\, $\downarrow$}  & {Ratio\,$\downarrow$} & {Valid\,$\uparrow$} & {Unique\, $\uparrow$} & {Novel\,$\uparrow$} & {Val., Uniq. \& Nov.\, $\uparrow$} & {$t$ (s)\,$\downarrow$}\\
\midrule
{Training set} & {0.0002} & {0.0310} & {0.0005} & {0.0052} & {0.0012} & {1.0} & {100.0} & {100.0} & {---} & {---} & {---}\\
\midrule
{GraphRNN}      & {0.0049} & {0.2779} & {1.2543} & {0.0459} & {0.1034} & {527.4} & {0.0} & {100.0} & {100.0} & {0.0} & {0.774}\\
{GRAN}          & {0.0007} & {0.0426} & {0.0009} & {0.0075} & {0.0019} & {1.9} & {97.5} & {85.0} & {2.5} & {0.0} & {0.920}\\ 
\midrule
{MolGAN*}       & {0.0009} & {0.3164} & {1.1730} & {0.1989} & {0.0729} & {491.9} & {0.0} & {25.0} & {100.0} & {0.0} & {0.002}\\ 
{GG-GAN (RS)*}  & {0.1005} & {0.2571} & {1.0313} & {0.2040} & {0.3829} & {586.3} & {0.0} & {100.0} & {100.0} & {0.0} & {0.011}\\ 
{GG-GAN*}       & {0.0630} & {1.1820} & {1.2280} & {0.1990} & {0.1890} & {601.0} & {0.0} & {10.0} & {100.0} & {0.0} & {0.011}\\ 
\rowcolor{Gray}
{\name ($k=2$)} & {0.0005} & {0.0785} & {0.0012} & {0.0112} & {0.0059} & {2.9} & {25.0} & {100.0} & {100.0} & {25.0} & {0.026}\\ \midrule
\rowcolor{Gray}
{\name ($k=2$, real spectra)} & {0.0010} & {0.0668} & {0.0010} & {0.0095} & {0.0056} & {3.1} & {47.5} & {100.0$^\ddagger$} & {100.0$^\ddagger$} & {47.5$^\ddagger$} & {0.011}\\ 
\midrule

 & \multicolumn{11}{c}{Stochastic Block Model}\\
\cmidrule(lr){2-12}
{Model} & {Deg.\,$\downarrow$} & {Clus.\,$\downarrow$} & {Orbit\, $\downarrow$} & {Spec.\,$\downarrow$} & {Wavelet\, $\downarrow$}  & {Ratio\,$\downarrow$} & {Valid\,$\uparrow$} & {Unique\, $\uparrow$} & {Novel\,$\uparrow$} & {Val., Uniq. \& Nov.\, $\uparrow$} & {$t$ (s)\,$\downarrow$}\\
\midrule
{Training set}  & {0.0008} & {0.0332} & {0.0255} & {0.0063} & {0.0007} & {1.0} & {100.0} & {100.0} & {---} & {---} & {---}\\
\midrule
{GraphRNN}      & {0.0055} & {0.0584} & {0.0785} & {0.0065} & {0.0431} & {14.9} & {5.0} & {100.0} & {100.0} & {5.0} & {5.108}\\ 
{GRAN}          & {0.0113} & {0.0553} & {0.0540} & {0.0054} & {0.0212} & {9.8} & {25.0} & {100.0} & {100.0} & {25.0} & {1.887}\\ \midrule
{MolGAN*}       & {0.0235} & {0.1161} & {0.0712} & {0.0117} & {0.0292} & {15.8} & {10.0} & {95.0} & {100.0} & {9.5} & {0.002} \\
{GG-GAN (RS)*}  & {0.0338} & {0.0581} & {0.1019} & {0.0613} & {0.1749} & {61.5} & {0.0} & {100.0} & {100.0} & {0.0} & {0.056} \\ {GG-GAN*}       & {0.0035} & {0.0699} & {0.0587} & {0.0094} & {0.0202} & {7.8} & {25.0} & {100.0} & {100.0} & {25.0} & {0.057} \\ 
\rowcolor{Gray}
{\name ($k=4$)} & {0.0015} & {0.0521} & {0.0412} & {0.0056} & {0.0028} & {2.0} & {52.5} & {100.0} & {100.0} & {52.5} & {0.074} \\ 
\midrule
\rowcolor{Gray}
{\name ($k=4$, real spectra)} & {0.0079} & {0.0528} & {0.0643} & {0.0074} & {0.0112} & {6.2} & {60.0} & {100.0$^\ddagger$} & {100.0$^\ddagger$} & {60.0$^\ddagger$} & {0.057} \\ 

\bottomrule
\end{tabular}}
\caption{Sample quality on synthetic datasets. We also provide \name results when the graph generator is conditioned on real spectra $(\blambda_k,\bU_k)$ from the test set. $^\ddagger$ novelty is compared to the test set graphs, from which $(\blambda_k,\bU_k)$ were taken. Methods marked with * are similar models implemented with building blocks from our architecture. 
}
\label{tab:sample_quality}
\end{table*}  

\paragraph{MMD measures.} Generated graph quality is commonly evaluated by comparing the distributions of graph statistics between the generated and test graphs \citep{li2018learning,you2018graphrnn,liao2019efficient,krawczuk2020gg}. In particular, we adopt the Maximum Mean Discrepancy (MMD) measures used by \citet{liao2019efficient}: node degree (Deg.), clustering coefficient (Clus.), orbit count (Orbit), eigenvalues of the normalized graph Laplacian (Spec.). We further introduce an eigenspace-based MMD (Wavelet) that evaluates the similarity of graph eigenspaces using statistics from a graph wavelet transform~\cite{hammond2011wavelets}. Similar features have been used to identify the role of each node on the graph \citep{donnat2018learning} --- see Appendix~\ref{app:spectral-mmd} for definition. The distance between the samples is computed using the total variational Gaussian kernel, which is consistent with the Gaussian earth mover's distance kernel while being considerably faster~\citep{liao2019efficient}. To stay consistent with published results for the Community dataset for it we use a Gaussian EMD kernel \cite{liu2019graph,pmlr-v108-niu20a}. We also summarize these measures by reporting the average ratio (Ratio) between a generator's and the training set's MMD values. As the training set MMD values are the best we can hope to achieve, this ratio indicates how far away we are from the optimal statistics.

\paragraph{Validity.} While the measures based on graph statistics measure how similar certain properties are between two distributions of graphs, they cannot guarantee that the generated graphs come from the same distribution. To this end, when the true distribution is known we also report a validity measure counting the percentage of generated graphs that belong to the ground-truth class. Specifically, we assert that planar graphs must be planar and connected and that SBM graphs are statistically indistinguishable from those generated by the ground-truth model (see Appendix~\ref{app:datasets}).

\paragraph{Novelty and uniqueness.}
While generating correct and statistically similar graphs is important, ultimately, a generator is useless unless the generated graphs are sufficiently diverse.
To ensure that our model generates sufficiently diverse graphs, we follow \citet{krawczuk2020gg} and introduce two measures based on graph isomorphism classes \citep{cordella2001improved}: 
\textit{Uniqueness.} We count the fraction of the generated graphs belonging to a unique isomorphism class. This way we check if the generator does not collapse to generating different permutations of the same few graphs.
\textit{Novelty.} We further determine what fraction of the generated graphs belong to an isomorphism class unseen in the training set. This measure further estimates how many novel graphs the generator is able to produce.

The validity, uniqueness, and novelty measures are summarized by computing the percentage of graphs that are all valid (if applicable), unique and novel. Together with the MMD ratio, this constitutes the two most important evaluation metrics we use.

\begin{table*}[ht]
\centering
\resizebox{1.0\linewidth}{!}{\begin{tabular}{l*{5}{S}g*{2}{S}gS}
\toprule
 & \multicolumn{10}{c}{Proteins}\\
\cmidrule(lr){2-11}
{Model} & {Deg.\,$\downarrow$} & {Clus.\,$\downarrow$} & {Orbit\,$\downarrow$} & {Spectral\,$\downarrow$} & {Wavelet\,$\downarrow$} & {Ratio\,$\downarrow$} & {Unique\,$\uparrow$} & {Novel\,$\uparrow$} & {Uniq. \& Nov.\,$\uparrow$} & {$t$ (s)\,$\downarrow$}\\
\midrule  
{Training set} & {0.0003} & {0.0068} & {0.0032} & {0.0009} & {0.0003} & {1.0} & {100.0} & {---} & {---} & {---}\\
\midrule  
{GraphRNN} & {0.0040} & {0.1475} & {0.5851} & {0.0152} & {0.0530} & {82.3} & {100.0} & {100.0} & {100.0} & {36.41}\\
{GRAN} & {0.0479} & {0.1234} & {0.3458} & {0.0125} & {0.0341} & {82.7} & {100.0} & {100.0} & {100.0} & {11.68}\\
\midrule  
{MolGAN*} & {0.0008} & {0.0644} & {0.0081} & {0.0021} & {0.0012} & {4.2} & {97.3} & {100.0} & {97.3} & {0.003}\\ 
{GG-GAN (RS)*} & {0.4727} & {0.1772} & {0.7326} & {0.4102} & {0.6278} & {875.8} & {100.0} & {100.0} & {100.0} & {0.482}\\
{GG-GAN*} & {0.5192} & {0.5220} & {0.7326} & {0.3996} & {0.6157} & {906.5} & {100.0} & {100.0} & {100.0} & {0.485}\\
\rowcolor{Gray}
{\name ($k=16$)} & {0.0056} & {0.0843} & {0.0267} & {0.0052} & {0.0118} & {16.9} & {100.0} & {100.0} & {100.0} & {0.507}\\
\midrule  
\rowcolor{Gray}
{\name ($k=16$, real spectra)} & {0.0013} & {0.0469} & {0.0287} & {0.0020} & {0.0022} & {6.0} & {100.0} & {100.0$^\ddagger$} & {100.0$^\ddagger$} & {0.485}\\ 

\bottomrule
\end{tabular}}
\caption{Protein graphs with up to 500 nodes. We also provide \name results when the graph generator is conditioned on real spectra $(\blambda_k,\bU_k)$ from the test set. $^\ddagger$ novelty is compared to the test set graphs, from which $(\blambda_k,\bU_k)$ were taken. Methods marked with * are similar models implemented with building blocks from our architecture.}
\label{tab:proteins}
\end{table*}  

\paragraph{Baselines.}
Besides the autoregressive GRAN and GraphRNN models, we construct GAN baselines inspired by MolGAN~\cite{de2018molgan} and GG-GAN \cite{krawczuk2020gg}. Neither of these baselines uses spectral conditioning and serve to prove the value of our approach. We choose to modify our architecture to create the baselines instead of faithfully re-implementing them, to ensure that improvements in score seen by \name do not come solely from other improvements we have made to the architecture, which include using a more powerful GNN, improved noise pre-processing and a symmetric architecture which is more stable to train. We create MolGAN*, by replacing our conditional graph generator with a three-layer MLP and a set of linear layers that convert final embedding to an adjacency matrix, node, and edge features (if needed). We retain the noise pre-processing MLP. We also build two models which use our PPGN-based generator. GG-GAN* uses learned fixed node embeddings alongside processed noise $\bw_A$ as generator input, while GG-GAN (RS)* depends only on the random pre-processed set $\bw_A$. In all cases, the discriminator architecture remains unchanged.

To compare how expensive it is to generate graphs with \name and with these baselines, we also provide the time it takes to generate one mini-batch of 10 graphs.

\paragraph{Model selection.}
GANs tend to oscillate around the optima. To this end, we track the exponential moving average of model weights with a retention coefficient of $0.995$. We also compute the MMD measures between training and validation sets and select the model with the best average ratio between its and the training set's MMD values.

\subsection{Results}

\paragraph{Synthetic datasets.} Table~\ref{tab:sample_quality} reports our results for distributions consisting of planar and SBM graphs and shows how \name, respectively, achieves a 169.6$\times$ and 3.9$\times$ improvement over the best non-overfitting baseline. Examples of generated graphs can be found in Figure~\ref{fig:graph_samples} and Appendix~\ref{appx:samples}. We see that only \name is able to produce novel and valid planar graphs. This example also highlights the importance of measuring generated graph uniqueness, as GRAN is able to produce perfectly valid graphs, but those graphs are memorized from the training set. When looking at SBM graphs, we observe that \name achieves much better MMD scores than alternatives.
It is likely that the diverse number of nodes and communities causes GRAN to underperform here, while the constant number of nodes in planar graphs helps it overfit.
Notice also how GAN baselines that do not utilize spectral conditioning do not produce good results on these larger graphs, whereas they work much better when the number of nodes is reduced (see e.g., the Community results in Table~\ref{tab:community}). The latter confirms our supposition that using spectral conditioning to separate global and local scale graph structure generation helps to overcome the expressivity limits of one-shot GAN generators.

\paragraph{Proteins.}
As seen in Table~\ref{tab:proteins}, the findings made on synthetic datasets also hold to larger real-world protein graphs. Here, the expressivity issues are even more pronounced, with some GAN baselines mostly failing to train. A notable exception is MolGAN*, which, in part due to our improved discriminator, is able to produce unique graphs with great statistical measures. Nevertheless, upon closer inspection, we observe that the graphs generated by MolGAN* have the tendency to be minor variations of a small number of graphs. In fact, on average only $17.6\%$ of edges differ between any two graphs produced by MolGAN* (Appendix \ref{appx:molgan_adj}).
Using teacher forcing and conditioning the generator and the discriminator on true spectral features in \name helps to avoid mode-collapse, which is a common issue with GANs, and helps to guide the model to convergence. Figure~\ref{fig:graph_samples} and Appendix~\ref{appx:samples} display generated graphs.

\paragraph{Conditioning on real spectra.}
In Tables~\ref{tab:sample_quality} and~\ref{tab:proteins}, we additionally provide the results achieved by the same \name model, when conditioned on \emph{true} first $k$ eigenvalues and eigenvectors taken form test graphs. 
We observe that when \name fails to capture the correct graph distribution (e.g. high MMDs in Table \ref{tab:proteins}), conditioning on real spectra significantly improves the generation process. 
This behavior suggests that there is still room for improvement in our spectrum generation procedure, especially when generating a large number of eigenvectors and eigenvalues. Furthermore, it is a sign that the conditional graph generator is able to leverage accurate spectral features without overfitting. In contrast, in the case when \name achieves a very good fit of the training distribution (Table \ref{tab:sample_quality}),  the percentage of valid graphs (i.e., graphs that are planar or match the true SBM parameters) improves when using true spectra while some MMD measures, most notably the degree MMD, become worse. 
This signals slight overfitting on the generated spectra by the graph generator $g_A$ and some divergence between the generated and true spectra. Note that training $g_A$ only on the true spectra would likely improve results.

\begin{table}[t]
\centering
\resizebox{1.0\columnwidth}{!}{\begin{tabular}{l*{3}{S}}
\toprule
{Dataset} & {Eigenvalue} & {Wavelet (true)} & {Wavelet (fake)} \\ 
\midrule  
{Planar (k=2)} & {17.60} & {45.43} & {206.15}\\
{SBM (k=4)} & {9.39} & {7.01}  & {19.46}\\
{Proteins (k=16)} & {36.83} & {4.94} & {23.42}\\
\bottomrule
\end{tabular}}
\caption{MMD ratios (vs training set MMD) for generated $k$ eigenvalues (direct MMD) and generated eigenvectors (Wavelet MMD) conditioned on fake and true test eigenvalues.}
\label{tab:spectral_mmd}
\end{table}

In Table \ref{tab:spectral_mmd}, we test the quality of the $k$ generated eigenvalues and eigenvectors using respectively a direct MMD for the eigenvalues and the Wavelet MMD (See~\ref{app:spectral-mmd}) for the eigenvectors. We evaluate the eigenvector generator both on the true and on the generated set of eigenvalues. As the overall MMD ratios are high, it indicates that a) the spectral generation procedure can be improved, and that b) conditioning even on imperfect spectra can significantly improve one-shot graph generation.

\paragraph{Smaller datasets.}
\begin{table}[t]
\centering
\resizebox{0.85\columnwidth}{!}{
\begin{tabular}{l*{3}{S}g}
\toprule
 & \multicolumn{4}{c}{Community-small}\\
\cmidrule(lr){2-5}
{Model} & {Deg.\,$\downarrow$} & {Clus.\,$\downarrow$} & {Orbit\,$\downarrow$} & {Ratio\,$\downarrow$}\\
\midrule  
{Training set} & {0.02} & {0.07} & {0.01} & {1.0} \\
\midrule  
{GraphRNN} & {0.08} & {0.12} & {0.04} & {3.2} \\
{GRAN} & {0.06} & {0.11} & {0.01} & {1.9} \\
\midrule  
{GraphVAE} & {0.35} & {0.98} & {0.54} & {28.5} \\
{DeepGMG} & {0.22} & {0.95} & {0.40} & {21.5} \\
{GNF} & {0.02} & {0.20} & {0.17} & {6.9} \\
{EDP-GNN} & {0.05} & {0.14} & {0.03} & {2.5} \\
\midrule  
{MolGAN*} & {0.06} & {0.13} & {0.01}  & {1.9} \\
{GG-GAN (RS)*} & {0.04} & {0.53} & {0.03} & {4.9} \\
{GG-GAN*} & {0.08} & {0.22} & {0.08} & {5.5} \\
\rowcolor{Gray}
{\name ($k=2$)} & {0.02} & {0.21} & {0.01} & {1.7} \\ 
\bottomrule
\end{tabular}}
\caption{Community graphs with up to 20 nodes. Methods marked with * are similar models implemented with building blocks from our architecture.
Other baseline results are taken from \cite{liu2019graph,pmlr-v108-niu20a}}
\label{tab:community}
\end{table}  
We also compared our method to a larger number of 
baselines that are unable to successfully generate larger graphs on two standard datasets consisting of small community graphs and molecules.
As shown in Table~\ref{tab:community}, \name achieves the smallest average MMD ratio.
While the MLP generator (MolGAN*) performs well on this simple graph distribution, we see that spectral conditioning considerably improves performance as compared to other equivariant GANs. 

\begin{table}[t]
\centering
\resizebox{1\columnwidth}{!}{\begin{tabular}{l*{2}{S}g}
\toprule
 & \multicolumn{3}{c}{QM9}\\
\cmidrule(lr){2-4}
{Model} & {Valid\,$\uparrow$} & {Val. \& Uniq.\,$\uparrow$} & {Val., Uniq. \& Nov.\,$\uparrow$}\\
\midrule  
{CharacterVAE} & {10.3} & {7.0} & {6.3} \\
{GrammarVAE} & {60.2} & {5.6} & {4.5} \\
{GraphVAE}  & {55.7} & {42.0} & {26.1} \\
{GraphVAE NoGM} & {81.0} & {20.5} & {11.9} \\
{GraphTransformerVAE}  & {74.6} & {16.8} & {15.8} \\
{MolGAN}  & {98.1} & {10.2} & {9.6} \\
\midrule  
{MolGAN*}  & {99.0} & {0.6} & {0.6} \\
{GG-GAN (RS)*}  & {51.2} & {24.4} & {24.4} \\
{GG-GAN* ($|\text{emb}| = 64$)}  & {100.0} & {0.4} & {0.4} \\
{GG-GAN* ($|\text{emb}| = 2$)}  & {6.6} & {1.8} & {1.8} \\
\rowcolor{Gray}
{\name ($k=2$)} & {87.3} & {31.2} & {29.1} \\
\bottomrule
\end{tabular}}
\caption{QM9 molecules with up to 9 heavy atoms. Methods marked with * are similar models implemented with building blocks from our architecture. Other baseline results are taken from \citep{mitton2021graph,de2018molgan}.\protect\footnotemark}
\label{tab:qm9}
\end{table}  

\footnotetext{Note, that we used the evaluation code provided by \cite{vignac2021top, de2018molgan}, which accepts samples with multiple connected components. If we instead only keep the largest connected component \name  Val., Uniq. \& Nov. score drops to 19.1.}

Table~\ref{tab:qm9} presents our results on the QM9 molecule dataset, which also involves node and edge feature generation. 
As highlighted by~\citet{de2018molgan}, GANs are prone to mode collapse on this dataset. We found that our strong PPGN discriminator aggravates this problem, especially for the GG-GAN* and MolGAN* variants.\footnote{We attempted to prevent mode collapse for our GG-GAN* and MolGAN* by increasing dropout rates, increasing the perturbations used for gradient penalty, and, removing the noise pre-processing MLP. None of these helped, besides reducing the number of dimensions used by the learned embeddings for GG-GAN*.} On the other hand, the GG-GAN (RS)* and \name methods proved more robust, which again highlights the power of conditional spectral generation in preventing mode collapse.

\section{Conclusion}


We argue that spectral conditioning is a key step in overcoming the challenges of one-shot graph generation: it prevents mode collapse, thus favoring the generation of novel and unique graphs, and it aids the generator to control the global structure leading to higher quality samples.
Our experiments show that \name outperforms competing methods with respect to generation time and sample quality, while also tending to generate more unique, valid, and novel graphs.

\section*{Disclosure of Funding}
Andreas Loukas would like to thank the Swiss National Science Foundation for supporting him in the context of the project “Deep Learning for Graph Structured Data”, grant number PZ00P2 179981. 

\bibliography{our.bib}

\begin{thebibliography}{81}
\providecommand{\natexlab}[1]{#1}
\providecommand{\url}[1]{\texttt{#1}}
\expandafter\ifx\csname urlstyle\endcsname\relax
  \providecommand{\doi}[1]{doi: #1}\else
  \providecommand{\doi}{doi: \begingroup \urlstyle{rm}\Url}\fi

\bibitem[Alon(1986)]{alon1986eigenvalues}
Alon, N.
\newblock Eigenvalues and expanders.
\newblock \emph{Combinatorica}, 6\penalty0 (2):\penalty0 83--96, 1986.

\bibitem[Alon \& Milman(1985)Alon and Milman]{alon1985lambda1}
Alon, N. and Milman, V.~D.
\newblock $\lambda$1, isoperimetric inequalities for graphs, and
  superconcentrators.
\newblock \emph{Journal of Combinatorial Theory, Series B}, 38\penalty0
  (1):\penalty0 73--88, 1985.

\bibitem[Arjovsky et~al.(2017)Arjovsky, Chintala, and
  Bottou]{arjovsky2017wasserstein}
Arjovsky, M., Chintala, S., and Bottou, L.
\newblock Wasserstein generative adversarial networks.
\newblock In \emph{International Conference on Machine Learning}, pp.\
  214--223, 2017.

\bibitem[Ba et~al.(2016)Ba, Kiros, and Hinton]{ba2016layer}
Ba, J.~L., Kiros, J.~R., and Hinton, G.~E.
\newblock Layer normalization.
\newblock \emph{Advances in NIPS 2016 Deep Learning Symposium}, 2016.

\bibitem[Baldesi et~al.(2018)Baldesi, Butts, and
  Markopoulou]{baldesi2018spectral}
Baldesi, L., Butts, C.~T., and Markopoulou, A.
\newblock Spectral graph forge: Graph generation targeting modularity.
\newblock In \emph{IEEE INFOCOM 2018-IEEE Conference on Computer
  Communications}, pp.\  1727--1735. IEEE, 2018.

\bibitem[Belkin \& Niyogi(2003)Belkin and Niyogi]{belkin2003laplacian}
Belkin, M. and Niyogi, P.
\newblock Laplacian eigenmaps for dimensionality reduction and data
  representation.
\newblock \emph{Neural computation}, 15\penalty0 (6):\penalty0 1373--1396,
  2003.

\bibitem[Bevilacqua et~al.(2021)Bevilacqua, Frasca, Lim, Srinivasan, Cai,
  Balamurugan, Bronstein, and Maron]{bevilacqua2021equivariant}
Bevilacqua, B., Frasca, F., Lim, D., Srinivasan, B., Cai, C., Balamurugan, G.,
  Bronstein, M.~M., and Maron, H.
\newblock Equivariant subgraph aggregation networks.
\newblock \emph{arXiv preprint arXiv:2110.02910}, 2021.

\bibitem[Bojchevski et~al.(2018)Bojchevski, Shchur, Z{\"{u}}gner, and
  G{\"{u}}nnemann]{bojchevski2018netgan}
Bojchevski, A., Shchur, O., Z{\"{u}}gner, D., and G{\"{u}}nnemann, S.
\newblock Netgan: Generating graphs via random walks.
\newblock In \emph{Proceedings of the 35th International Conference on Machine
  Learning, {ICML} 2018, Stockholmsm{\"{a}}ssan, Stockholm, Sweden, July 10-15,
  2018}, pp.\  609--618, 2018.

\bibitem[Bouritsas et~al.(2020)Bouritsas, Frasca, Zafeiriou, and
  Bronstein]{bouritsas2020improving}
Bouritsas, G., Frasca, F., Zafeiriou, S., and Bronstein, M.~M.
\newblock Improving graph neural network expressivity via subgraph isomorphism
  counting.
\newblock \emph{arXiv preprint arXiv:2006.09252}, 2020.

\bibitem[Bresson \& Laurent(2019)Bresson and Laurent]{bresson2019two}
Bresson, X. and Laurent, T.
\newblock A two-step graph convolutional decoder for molecule generation.
\newblock In \emph{NeurIPS Workshop on Machine Learning and the Physical
  Sciences}, 2019.

\bibitem[Brock et~al.(2018)Brock, Donahue, and Simonyan]{brock2018large}
Brock, A., Donahue, J., and Simonyan, K.
\newblock Large scale gan training for high fidelity natural image synthesis.
\newblock In \emph{International Conference on Learning Representations}, 2018.

\bibitem[Chakrabarti \& Faloutsos(2006)Chakrabarti and
  Faloutsos]{chakrabarti2006graph}
Chakrabarti, D. and Faloutsos, C.
\newblock Graph mining: Laws, generators, and algorithms.
\newblock \emph{ACM computing surveys (CSUR)}, 38\penalty0 (1):\penalty0 2--es,
  2006.

\bibitem[Chung \& Graham(1997)Chung and Graham]{chung1997spectral}
Chung, F.~R. and Graham, F.~C.
\newblock \emph{Spectral graph theory}.
\newblock Number~92. American Mathematical Soc., 1997.

\bibitem[Coifman et~al.(2005)Coifman, Lafon, Lee, Maggioni, Nadler, Warner, and
  Zucker]{coifman2005geometric}
Coifman, R.~R., Lafon, S., Lee, A.~B., Maggioni, M., Nadler, B., Warner, F.,
  and Zucker, S.~W.
\newblock Geometric diffusions as a tool for harmonic analysis and structure
  definition of data: Diffusion maps.
\newblock \emph{Proceedings of the national academy of sciences}, 102\penalty0
  (21):\penalty0 7426--7431, 2005.

\bibitem[Cordella et~al.(2001)Cordella, Foggia, Sansone, and
  Vento]{cordella2001improved}
Cordella, L.~P., Foggia, P., Sansone, C., and Vento, M.
\newblock An improved algorithm for matching large graphs.
\newblock In \emph{3rd IAPR-TC15 workshop on graph-based representations in
  pattern recognition}, pp.\  149--159. Citeseer, 2001.

\bibitem[Dai et~al.(2020)Dai, Nazi, Li, Dai, and Schuurmans]{dai2020scalable}
Dai, H., Nazi, A., Li, Y., Dai, B., and Schuurmans, D.
\newblock Scalable deep generative modeling for sparse graphs.
\newblock In \emph{International Conference on Machine Learning}, pp.\
  2302--2312. PMLR, 2020.

\bibitem[De~Cao \& Kipf(2018)De~Cao and Kipf]{de2018molgan}
De~Cao, N. and Kipf, T.
\newblock Molgan: An implicit generative model for small molecular graphs.
\newblock \emph{arXiv preprint arXiv:1805.11973}, 2018.

\bibitem[Defferrard et~al.()Defferrard, Martin, Pena, and Perraudin]{pygsp}
Defferrard, M., Martin, L., Pena, R., and Perraudin, N.
\newblock Pygsp: Graph signal processing in python.

\bibitem[Dobson \& Doig(2003)Dobson and Doig]{dobson2003distinguishing}
Dobson, P.~D. and Doig, A.~J.
\newblock Distinguishing enzyme structures from non-enzymes without alignments.
\newblock \emph{Journal of molecular biology}, 330\penalty0 (4):\penalty0
  771--783, 2003.

\bibitem[Donahue et~al.(2018)Donahue, McAuley, and
  Puckette]{donahue2018adversarial}
Donahue, C., McAuley, J., and Puckette, M.
\newblock Adversarial audio synthesis.
\newblock In \emph{International Conference on Learning Representations}, 2018.

\bibitem[Donnat et~al.(2018)Donnat, Zitnik, Hallac, and
  Leskovec]{donnat2018learning}
Donnat, C., Zitnik, M., Hallac, D., and Leskovec, J.
\newblock Learning structural node embeddings via diffusion wavelets.
\newblock In \emph{Proceedings of the 24th ACM SIGKDD International Conference
  on Knowledge Discovery \& Data Mining}, pp.\  1320--1329, 2018.

\bibitem[Edelman et~al.(1998)Edelman, Arias, and Smith]{edelman1998geometry}
Edelman, A., Arias, T.~A., and Smith, S.~T.
\newblock The geometry of algorithms with orthogonality constraints.
\newblock \emph{SIAM journal on Matrix Analysis and Applications}, 20\penalty0
  (2):\penalty0 303--353, 1998.

\bibitem[Eldridge et~al.(2016)Eldridge, Belkin, and Wang]{eldridge2016graphons}
Eldridge, J., Belkin, M., and Wang, Y.
\newblock Graphons, mergeons, and so on!
\newblock In \emph{Advances in Neural Information Processing Systems}, pp.\
  2307--2315, 2016.

\bibitem[Erdos et~al.(1960)Erdos, R{\'e}nyi, et~al.]{erdos1960evolution}
Erdos, P., R{\'e}nyi, A., et~al.
\newblock On the evolution of random graphs.
\newblock \emph{Publ. Math. Inst. Hung. Acad. Sci}, 5\penalty0 (1):\penalty0
  17--60, 1960.

\bibitem[Fahrmeir et~al.(2007)Fahrmeir, Kneib, Lang, and
  Marx]{fahrmeir2007regression}
Fahrmeir, L., Kneib, T., Lang, S., and Marx, B.
\newblock \emph{Regression}.
\newblock Springer, 2007.

\bibitem[Garcia~Satorras et~al.(2021)Garcia~Satorras, Hoogeboom, Fuchs, Posner,
  and Welling]{garcia2021n}
Garcia~Satorras, V., Hoogeboom, E., Fuchs, F., Posner, I., and Welling, M.
\newblock E (n) equivariant normalizing flows.
\newblock \emph{Advances in Neural Information Processing Systems}, 34, 2021.

\bibitem[Goldenberg et~al.(2010)Goldenberg, Zheng, Fienberg, and
  Airoldi]{goldenberg2010survey}
Goldenberg, A., Zheng, A.~X., Fienberg, S.~E., and Airoldi, E.~M.
\newblock A survey of statistical network models.
\newblock 2010.

\bibitem[Goodfellow et~al.(2014)Goodfellow, Pouget-Abadie, Mirza, Xu,
  Warde-Farley, Ozair, Courville, and Bengio]{goodfellow2014generative}
Goodfellow, I., Pouget-Abadie, J., Mirza, M., Xu, B., Warde-Farley, D., Ozair,
  S., Courville, A., and Bengio, Y.
\newblock Generative adversarial nets.
\newblock \emph{Advances in neural information processing systems}, 27, 2014.

\bibitem[Gulrajani et~al.(2017)Gulrajani, Ahmed, Arjovsky, Dumoulin, and
  Courville]{gulrajani2017improved}
Gulrajani, I., Ahmed, F., Arjovsky, M., Dumoulin, V., and Courville, A.
\newblock Improved training of wasserstein gans, 2017.

\bibitem[Hammond et~al.(2011)Hammond, Vandergheynst, and
  Gribonval]{hammond2011wavelets}
Hammond, D.~K., Vandergheynst, P., and Gribonval, R.
\newblock Wavelets on graphs via spectral graph theory.
\newblock \emph{Applied and Computational Harmonic Analysis}, 30\penalty0
  (2):\penalty0 129--150, 2011.

\bibitem[Hammond et~al.(2013)Hammond, Gur, and Johnson]{hammond2013graph}
Hammond, D.~K., Gur, Y., and Johnson, C.~R.
\newblock Graph diffusion distance: A difference measure for weighted graphs
  based on the graph laplacian exponential kernel.
\newblock In \emph{2013 IEEE Global Conference on Signal and Information
  Processing}, pp.\  419--422. IEEE, 2013.

\bibitem[Hendrycks \& Gimpel(2016)Hendrycks and Gimpel]{hendrycks2016gaussian}
Hendrycks, D. and Gimpel, K.
\newblock Gaussian error linear units (gelus).
\newblock \emph{arXiv preprint arXiv:1606.08415}, 2016.

\bibitem[Holland et~al.(1983)Holland, Laskey, and
  Leinhardt]{holland1983stochastic}
Holland, P.~W., Laskey, K.~B., and Leinhardt, S.
\newblock Stochastic blockmodels: First steps.
\newblock \emph{Social networks}, 5\penalty0 (2):\penalty0 109--137, 1983.

\bibitem[Huang et~al.(2016)Huang, Boyken, and Baker]{huang2016coming}
Huang, P.-S., Boyken, S.~E., and Baker, D.
\newblock The coming of age of de novo protein design.
\newblock \emph{Nature}, 537\penalty0 (7620):\penalty0 320--327, 2016.

\bibitem[Jang et~al.(2016)Jang, Gu, and Poole]{jang2016categorical}
Jang, E., Gu, S., and Poole, B.
\newblock Categorical reparameterization with gumbel-softmax.
\newblock 2016.

\bibitem[Jin et~al.(2018)Jin, Barzilay, and Jaakkola]{jin2018junction}
Jin, W., Barzilay, R., and Jaakkola, T.
\newblock Junction tree variational autoencoder for molecular graph generation.
\newblock In \emph{International conference on machine learning}, pp.\
  2323--2332. PMLR, 2018.

\bibitem[Jin et~al.(2020)Jin, Barzilay, and Jaakkola]{pmlr-v119-jin20a}
Jin, W., Barzilay, D., and Jaakkola, T.
\newblock Hierarchical generation of molecular graphs using structural motifs.
\newblock In \emph{Proceedings of the 37th International Conference on Machine
  Learning}, volume 119 of \emph{Proceedings of Machine Learning Research},
  pp.\  4839--4848. PMLR, 13--18 Jul 2020.

\bibitem[Kingma \& Ba(2015)Kingma and Ba]{kingma2014adam}
Kingma, D.~P. and Ba, J.
\newblock Adam: {A} method for stochastic optimization.
\newblock In \emph{3rd International Conference on Learning Representations,
  {ICLR} 2015, San Diego, CA, USA, May 7-9, 2015, Conference Track
  Proceedings}, 2015.

\bibitem[Kipf \& Welling(2016)Kipf and Welling]{kipf2016variational}
Kipf, T.~N. and Welling, M.
\newblock Variational graph auto-encoders.
\newblock In \emph{NIPS Workshop onBayesian Deep Learning}, 2016.

\bibitem[Krawczuk et~al.(2020)Krawczuk, Abranches, Loukas, and
  Cevher]{krawczuk2020gg}
Krawczuk, I., Abranches, P., Loukas, A., and Cevher, V.
\newblock Gg-gan: A geometric graph generative adversarial network.
\newblock 2020.

\bibitem[Lee et~al.(2014)Lee, Gharan, and Trevisan]{lee2014multiway}
Lee, J.~R., Gharan, S.~O., and Trevisan, L.
\newblock Multiway spectral partitioning and higher-order cheeger inequalities.
\newblock \emph{Journal of the ACM (JACM)}, 61\penalty0 (6):\penalty0 1--30,
  2014.

\bibitem[Li et~al.(2018)Li, Vinyals, Dyer, Pascanu, and
  Battaglia]{li2018learning}
Li, Y., Vinyals, O., Dyer, C., Pascanu, R., and Battaglia, P.
\newblock Learning deep generative models of graphs.
\newblock 2018.

\bibitem[Liao et~al.(2019)Liao, Li, Song, Wang, Hamilton, Duvenaud, Urtasun,
  and Zemel]{liao2019efficient}
Liao, R., Li, Y., Song, Y., Wang, S., Hamilton, W., Duvenaud, D.~K., Urtasun,
  R., and Zemel, R.
\newblock Efficient graph generation with graph recurrent attention networks.
\newblock In \emph{Advances in Neural Information Processing Systems}, pp.\
  4255--4265, 2019.

\bibitem[Liu et~al.(2019)Liu, Kumar, Ba, Kiros, and Swersky]{liu2019graph}
Liu, J., Kumar, A., Ba, J., Kiros, J., and Swersky, K.
\newblock Graph normalizing flows.
\newblock \emph{Advances in Neural Information Processing Systems},
  32:\penalty0 13578--13588, 2019.

\bibitem[Liu et~al.(2021)Liu, Yan, Oztekin, and Ji]{liu2021graphebm}
Liu, M., Yan, K., Oztekin, B., and Ji, S.
\newblock Graph{EBM}: Molecular graph generation with energy-based models.
\newblock In \emph{Energy Based Models Workshop - ICLR 2021}, 2021.

\bibitem[Liu et~al.(2018)Liu, Allamanis, Brockschmidt, and
  Gaunt]{liu2018constrained}
Liu, Q., Allamanis, M., Brockschmidt, M., and Gaunt, A.~L.
\newblock Constrained graph variational autoencoders for molecule design.
\newblock In \emph{Proceedings of the 32nd International Conference on Neural
  Information Processing Systems}, pp.\  7806--7815, 2018.

\bibitem[Maddison et~al.(2017)Maddison, Mnih, and Teh]{maddison2016concrete}
Maddison, C., Mnih, A., and Teh, Y.
\newblock The concrete distribution: A continuous relaxation of discrete random
  variables.
\newblock In \emph{Proceedings of the international conference on learning
  Representations}. International Conference on Learning Representations, 2017.

\bibitem[Mahmood et~al.(2021)Mahmood, Mansimov, Bonneau, and
  Cho]{mahmood2021masked}
Mahmood, O., Mansimov, E., Bonneau, R., and Cho, K.
\newblock Masked graph modeling for molecule generation.
\newblock \emph{Nature communications}, 12\penalty0 (1):\penalty0 1--12, 2021.

\bibitem[Maron et~al.(2019)Maron, Ben-Hamu, Serviansky, and
  Lipman]{maron2019provably}
Maron, H., Ben-Hamu, H., Serviansky, H., and Lipman, Y.
\newblock Provably powerful graph networks.
\newblock In \emph{Advances in Neural Information Processing Systems}, pp.\
  2156--2167, 2019.

\bibitem[Mirhoseini et~al.(2021)Mirhoseini, Goldie, Yazgan, Jiang, Songhori,
  Wang, Lee, Johnson, Pathak, Nazi, et~al.]{mirhoseini2021graph}
Mirhoseini, A., Goldie, A., Yazgan, M., Jiang, J.~W., Songhori, E., Wang, S.,
  Lee, Y.-J., Johnson, E., Pathak, O., Nazi, A., et~al.
\newblock A graph placement methodology for fast chip design.
\newblock \emph{Nature}, 594\penalty0 (7862):\penalty0 207--212, 2021.

\bibitem[Mitton et~al.(2021)Mitton, Senn, Wynne, and
  Murray-Smith]{mitton2021graph}
Mitton, J., Senn, H.~M., Wynne, K., and Murray-Smith, R.
\newblock A graph {VAE} and graph transformer approach to generating molecular
  graphs.
\newblock In \emph{Graph Representation learning and beyong (ICLR Workshop)},
  2021.

\bibitem[Mohar(1997)]{mohar1997some}
Mohar, B.
\newblock Some applications of laplace eigenvalues of graphs.
\newblock In \emph{Graph symmetry}, pp.\  225--275. Springer, 1997.

\bibitem[Morris et~al.(2019)Morris, Ritzert, Fey, Hamilton, Lenssen, Rattan,
  and Grohe]{morris2019weisfeiler}
Morris, C., Ritzert, M., Fey, M., Hamilton, W.~L., Lenssen, J.~E., Rattan, G.,
  and Grohe, M.
\newblock Weisfeiler and leman go neural: Higher-order graph neural networks.
\newblock In \emph{Proceedings of the AAAI Conference on Artificial
  Intelligence}, volume~33, pp.\  4602--4609, 2019.

\bibitem[Morris et~al.(2020)Morris, Rattan, and Mutzel]{NEURIPS2020_f81dee42}
Morris, C., Rattan, G., and Mutzel, P.
\newblock Weisfeiler and leman go sparse: Towards scalable higher-order graph
  embeddings.
\newblock In \emph{Advances in Neural Information Processing Systems},
  volume~33, pp.\  21824--21840. Curran Associates, Inc., 2020.

\bibitem[Niu et~al.(2020)Niu, Song, Song, Zhao, Grover, and
  Ermon]{pmlr-v108-niu20a}
Niu, C., Song, Y., Song, J., Zhao, S., Grover, A., and Ermon, S.
\newblock Permutation invariant graph generation via score-based generative
  modeling.
\newblock volume 108 of \emph{Proceedings of Machine Learning Research}, pp.\
  4474--4484, Online, 26--28 Aug 2020. PMLR.

\bibitem[Oord et~al.(2016{\natexlab{a}})Oord, Dieleman, Zen, Simonyan, Vinyals,
  Graves, Kalchbrenner, Senior, and Kavukcuoglu]{oord2016wavenet}
Oord, A. v.~d., Dieleman, S., Zen, H., Simonyan, K., Vinyals, O., Graves, A.,
  Kalchbrenner, N., Senior, A., and Kavukcuoglu, K.
\newblock Wavenet: A generative model for raw audio.
\newblock \emph{arXiv preprint arXiv:1609.03499}, 2016{\natexlab{a}}.

\bibitem[Oord et~al.(2016{\natexlab{b}})Oord, Kalchbrenner, Vinyals, Espeholt,
  Graves, and Kavukcuoglu]{oord2016conditional}
Oord, A. v.~d., Kalchbrenner, N., Vinyals, O., Espeholt, L., Graves, A., and
  Kavukcuoglu, K.
\newblock Conditional image generation with pixelcnn decoders.
\newblock In \emph{Proceedings of the 30th International Conference on Neural
  Information Processing Systems}, NIPS'16, 2016{\natexlab{b}}.

\bibitem[Papp et~al.(2021)Papp, Martinkus, Faber, and
  Wattenhofer]{papp2021dropgnn}
Papp, P.~A., Martinkus, K., Faber, L., and Wattenhofer, R.
\newblock Drop{GNN}: Random dropouts increase the expressiveness of graph
  neural networks.
\newblock volume~34, pp.\  21997--22009, 2021.

\bibitem[Peixoto(2019)]{peixoto2019bayesian}
Peixoto, T.~P.
\newblock Bayesian stochastic blockmodeling.
\newblock \emph{Advances in network clustering and blockmodeling}, pp.\
  289--332, 2019.

\bibitem[Peixoto(2020)]{peixoto2020merge}
Peixoto, T.~P.
\newblock Merge-split markov chain monte carlo for community detection.
\newblock \emph{Physical Review E}, 102\penalty0 (1):\penalty0 012305, 2020.

\bibitem[Perraudin(2017)]{perraudin2017graph}
Perraudin, N.
\newblock \emph{Graph-based structures in data science: fundamental limits and
  applications to machine learning}.
\newblock PhD thesis, Ecole Polytechnique F{\'e}d{\'e}rale de Lausanne, 2017.

\bibitem[Petzka et~al.(2018)Petzka, Fischer, and
  Lukovnikov]{petzka2018regularization}
Petzka, H., Fischer, A., and Lukovnikov, D.
\newblock On the regularization of wasserstein gans.
\newblock In \emph{International Conference on Learning Representations}, 2018.

\bibitem[Qi et~al.(2017)Qi, Su, Mo, and Guibas]{qi2017pointnet}
Qi, C.~R., Su, H., Mo, K., and Guibas, L.~J.
\newblock Pointnet: Deep learning on point sets for 3d classification and
  segmentation.
\newblock In \emph{Proceedings of the IEEE conference on computer vision and
  pattern recognition}, pp.\  652--660, 2017.

\bibitem[Ramakrishnan et~al.(2014)Ramakrishnan, Dral, Rupp, and
  Von~Lilienfeld]{ramakrishnan2014quantum}
Ramakrishnan, R., Dral, P.~O., Rupp, M., and Von~Lilienfeld, O.~A.
\newblock Quantum chemistry structures and properties of 134 kilo molecules.
\newblock \emph{Scientific data}, 1\penalty0 (1):\penalty0 1--7, 2014.

\bibitem[Sandfelder et~al.(2021)Sandfelder, Vijayan, and
  Hamilton]{sandfelder2021ego}
Sandfelder, D., Vijayan, P., and Hamilton, W.~L.
\newblock Ego-gnns: Exploiting ego structures in graph neural networks.
\newblock In \emph{ICASSP 2021-2021 IEEE International Conference on Acoustics,
  Speech and Signal Processing (ICASSP)}, pp.\  8523--8527. IEEE, 2021.

\bibitem[Segol \& Lipman(2019)Segol and Lipman]{segol2019universal}
Segol, N. and Lipman, Y.
\newblock On universal equivariant set networks.
\newblock In \emph{International Conference on Learning Representations}, 2019.

\bibitem[Serviansky et~al.(2020)Serviansky, Segol, Shlomi, Cranmer, Gross,
  Maron, and Lipman]{serviansky2020set2graph}
Serviansky, H., Segol, N., Shlomi, J., Cranmer, K., Gross, E., Maron, H., and
  Lipman, Y.
\newblock Set2graph: Learning graphs from sets.
\newblock In \emph{Advances in Neural Information Processing Systems},
  volume~33, pp.\  22080--22091, 2020.

\bibitem[Shepard et~al.(2015)Shepard, Brozell, and
  Gidofalvi]{shepard2015representation}
Shepard, R., Brozell, S.~R., and Gidofalvi, G.
\newblock The representation and parametrization of orthogonal matrices.
\newblock \emph{The Journal of Physical Chemistry A}, 119\penalty0
  (28):\penalty0 7924--7939, 2015.

\bibitem[Shi et~al.(2020)Shi, Fan, and Kwok]{shi2020effective}
Shi, H., Fan, H., and Kwok, J.~T.
\newblock Effective decoding in graph auto-encoder using triadic closure.
\newblock In \emph{Proceedings of the AAAI Conference on Artificial
  Intelligence}, volume~34, pp.\  906--913, 2020.

\bibitem[Shi \& Malik(2000)Shi and Malik]{shi2000normalized}
Shi, J. and Malik, J.
\newblock Normalized cuts and image segmentation.
\newblock \emph{IEEE Transactions on pattern analysis and machine
  intelligence}, 22\penalty0 (8):\penalty0 888--905, 2000.

\bibitem[Shine \& Kempe(2019)Shine and Kempe]{10.1145/3308558.3313631}
Shine, A. and Kempe, D.
\newblock Generative graph models based on laplacian spectra?
\newblock In \emph{The World Wide Web Conference}, WWW '19, pp.\  1691–1701,
  2019.

\bibitem[Simonovsky \& Komodakis(2018)Simonovsky and
  Komodakis]{simonovsky2018graphvae}
Simonovsky, M. and Komodakis, N.
\newblock Graphvae: Towards generation of small graphs using variational
  autoencoders.
\newblock In \emph{International conference on artificial neural networks},
  pp.\  412--422. Springer, 2018.

\bibitem[Sinclair \& Jerrum(1989)Sinclair and Jerrum]{SINCLAIR198993}
Sinclair, A. and Jerrum, M.
\newblock Approximate counting, uniform generation and rapidly mixing markov
  chains.
\newblock \emph{Information and Computation}, 82\penalty0 (1):\penalty0
  93--133, 1989.
\newblock ISSN 0890-5401.
\newblock \doi{https://doi.org/10.1016/0890-5401(89)90067-9}.

\bibitem[Vignac \& Frossard(2021)Vignac and Frossard]{vignac2021top}
Vignac, C. and Frossard, P.
\newblock Top-n: Equivariant set and graph generation without exchangeability.
\newblock \emph{arXiv preprint arXiv:2110.02096}, 2021.

\bibitem[Vignac et~al.(2020)Vignac, Loukas, and Frossard]{vignac2020building}
Vignac, C., Loukas, A., and Frossard, P.
\newblock Building powerful and equivariant graph neural networks with
  structural message-passing.
\newblock In \emph{NeurIPS}, 2020.

\bibitem[Wang et~al.(2018)Wang, Wang, Wang, Zhao, Zhang, Zhang, Xie, and
  Guo]{wang2018graphgan}
Wang, H., Wang, J., Wang, J., Zhao, M., Zhang, W., Zhang, F., Xie, X., and Guo,
  M.
\newblock Graphgan: Graph representation learning with generative adversarial
  nets.
\newblock In \emph{Proceedings of the AAAI conference on artificial
  intelligence}, volume~32, 2018.

\bibitem[Xu et~al.(2019)Xu, Hu, Leskovec, and Jegelka]{xu2018how}
Xu, K., Hu, W., Leskovec, J., and Jegelka, S.
\newblock How powerful are graph neural networks?
\newblock In \emph{International Conference on Learning Representations}, 2019.

\bibitem[Yang et~al.(2019)Yang, Zhuang, Shi, Luu, and Li]{yang2019conditional}
Yang, C., Zhuang, P., Shi, W., Luu, A., and Li, P.
\newblock Conditional structure generation through graph variational generative
  adversarial nets.
\newblock In \emph{Advances in Neural Information Processing Systems}, pp.\
  1340--1351, 2019.

\bibitem[You et~al.(2018{\natexlab{a}})You, Liu, Ying, Pande, and
  Leskovec]{you2018graph}
You, J., Liu, B., Ying, R., Pande, V., and Leskovec, J.
\newblock Graph convolutional policy network for goal-directed molecular graph
  generation.
\newblock In \emph{Proceedings of the 32nd International Conference on Neural
  Information Processing Systems}, NIPS'18, 2018{\natexlab{a}}.

\bibitem[You et~al.(2018{\natexlab{b}})You, Ying, Ren, Hamilton, and
  Leskovec]{you2018graphrnn}
You, J., Ying, R., Ren, X., Hamilton, W.~L., and Leskovec, J.
\newblock Graphrnn: Generating realistic graphs with deep auto-regressive
  models.
\newblock In \emph{ICML}, 2018{\natexlab{b}}.

\bibitem[Zhang \& Li(2021)Zhang and Li]{zhang2021nested}
Zhang, M. and Li, P.
\newblock Nested graph neural networks.
\newblock \emph{Advances in Neural Information Processing Systems}, 34, 2021.

\end{thebibliography}
\bibliographystyle{icml2022}

\clearpage
\newpage
\appendix

\section{Model Implementation Details}
\label{appx:architecture}

\textit{In this section, we present additional information describing in details our general architecture presented in Figure~\ref{fig:architecture} left.}

The input noise used by each generator is sampled from a Gaussian distribution $z \sim \mathcal{N}(0,1)$, normalized to unit length and processed by a 4-layer MLP, with input, output, and hidden sizes of $100$. Three different MLPs are used, each for a different generator. Every linear layer, including ones in MLPs except the final model output layer use layer normalization~\cite{ba2016layer}. We use GELU~\cite{hendrycks2016gaussian} activation function throughout our architecture, except for the eigenvalue generator which uses gated activation units~\citep{oord2016conditional,oord2016wavenet}. In our model, \emph{we ignore the very first eigenvalue and eigenvector} as the first normalized Laplacian eigenvector only caries degree information.

Both PPGN models~\citep{maron2019provably} have 8 layers (3 layers for QM9 experiments) with hidden dimension of $64$, PointNetST models~\citep{segol2019universal} use 3-layer MLPs with base size of $32$ (see Section~\ref{appx:pointnetst}) and the CNN models have 4 convolutional layers $[5\times32, 9\times16, 17\times8, 25\times1]$ for the generator, with discriminator having same architecture in reverse. For each dataset, we select the number of spectral components $k$ to be at least $2$, but as small as is sufficient for the conditional graph generator to achieve good validity, while conditioned only on true spectral features. Our MolGAN* generator uses MLP of size $[128, 256, 512]$ following~\citep{de2018molgan}. For GG-GAN* we use a learned embedding size of $64$.

The models are trained on a machine with two 64 Core AMD EPYC 7742 CPUs, 500GB RAM, and eight 24GB GeForce RTX 3090 GPUs. All of the individual runs only use one GPU at a time, except protein generation with GAN models. There we use 4 GPUs to achieve an effective batch size of 4. For SBM experiments we use a batch size of 5, for community and planar graphs a batch size of 10, and for QM9 following~\citep{de2018molgan} we use a batch size of 128.

\subsection{Provably Powerful Graph Network (PPGN)}\label{appx:ppgn}

The PPGN layer \citep{maron2019provably} consists of four operations on channel-first input $\bX \in \mathbb{R}^{h \times n \times n}$. First, two versions of the matrix are built using two MLPs applied to each feature independently, i.e., repeated across the $n \times n$ dimensions.
$$\bM_1 = \text{MLP}_1(\bX) \in \mathbb{R}^{h_2 \times n \times n},$$ 
$$\bM_2 = \text{MLP}_2(\bX) \in \mathbb{R}^{h_2 \times n \times n}.$$
The $h_2$ matrices of size $n\times n$ are then multiplied as
$$\bM^{i} = \bM_1^{i} \bM_2^{i}.$$
Eventually, the output is obtained by concatenating\footnote{The concatenation operation is written $(\cdot \mathbin\Vert \cdot )$ } $\bM$  together with the original transformed by a third MLP:
$$\bX^{t+1} = \text{MLP}_3(\bX^{t} \mathbin\Vert \bM).$$

We use 8 layers, an embedding size of 64, and a skip connection from each layer (including inputs after the initial embedding linear layer) to the final readout. In the readout, a linear layer is applied on the concatenation of all of the layer outputs. For the generator, this is passed through an MLP to produce the adjacency matrix, while for the discriminator it is read out by applying an MLP over the sum of diagonal values and another MLP over the sum of off-diagonal values~\citep{maron2019provably}. Instance normalization is used after each PPGN layer. Disregarding the skip (concatenation) connection and instance normalization, our PPGN architecture matches the original implementation with one small, but important improvement. We normalize $\bM^{i} = \bM_1^{i} \bM_2^{i}$ by $1/\sqrt{n}$, to avoid growth of the gradient and value variance. 
We apply a dropout of $0.1$ before the final readout linear layer in PPGN.

Extending PPGN to generate node and edge features as well is simple. 
a) To produce the node features, a readout MLP that uses the concatenation of the diagonal elements from all of the layer outputs can be added.
b) To produce the edge features, we change the adjacency readout MLP to produce not just a single channel matrix, but a multi-channel one, and interpret the new channels as different edge types or features.
These node and edge features can then be fed into the discriminator.

\subsection{PointNetST}\label{appx:pointnetst}
A PointNetST layer~\citep{segol2019universal} is mean to process set inputs $\bX^{n \times h}$ and is comprised of three MLPs: $\text{MLP}_{\text{feat}}$, $\text{MLP}_{\text{agg}}$ and $\text{MLP}_{\text{cat}}$. All MLPs have different relative sizes, with respect their base hidden dimension $h$. $\text{MLP}_{\text{feat}}$ is small ($[h, h, h]$) and is used for input pre-processing:
$$
\bY = \text{MLP}_{\text{feat}}(\bX) \in \mathbb{R}^{n\times h}.
$$
$\text{MLP}_{\text{agg}}$ enlarges the embeddings ($[h, h *2, h*4]$) and prepares them for global aggregation:
$$
\bar{\bY} = \frac{1}{n} \sum_n \text{MLP}_{\text{agg}}(\bX) \in \mathbb{R}^{1 \times 4*h}.
$$
$\text{MLP}_{\text{cat}}$ of size ($[h*4, h*2, \text{output dimension}]$) concatenates individual set element features $\bY$ with the (repeated) global set feature vector $\bar{\bY}$ and produces the final output:
$$
\bX = \text{MLP}_{\text{cat}}(\bY \mathbin\Vert \bar{\bY})  \in \mathbb{R}^{n\times \text{output dimension}}.
$$

Whenever a global output vector is required in our architecture we apply another MLP on the mean-aggregated outputs of the PointNetST layer as discussed before. In such cases PointNetST output dimension is set to $h*4$ and this final MLP is of size $[h*2, h, \text{final output dimension}]$. When this readout MLP is used we apply a dropout of $0.1$ to its inputs.

\subsection{Dynamic Number of Nodes}
To handle graphs with a varying number of nodes we create a binary mask using the conditioning number of nodes. We apply such $n \times h$ sized masks before any global set operation, such as mean (we also correct the mean for the true number of nodes). We also apply $n \times n \times h$ sized mask before any graph convolution or pooling (we again correct mean pooling for the true number of nodes).
This allows us to use batched dense tensor operations with dynamic graph sizes in our model.

\subsection{Gradient Penalty}
To ensure good convergence of our GANs, we use Wasserstein loss~\cite{wang2018graphgan} with gradient penalty~\cite{gulrajani2017improved}. In particular, we use the less restrictive gradient penalty formulation~\cite{petzka2018regularization}:
$$
\text{gp} = \gamma(\max{\{0, ||\nabla d(\hat{\bx})|| - 1\}})^2 ,
$$
where $d$ is a discriminator and $\hat{\bx}$ are inputs on which gradient penalty is computed.
In general, $\hat{\bx}$ is created by interpolation between true and fake samples. While interpolation of eigenvalues is straightforward, it is challenging to interpolate between eigenvectors. To interpolate eigenvectors, we first apply a canonical set ordering, which is achieved by flipping the eigenvector sign, such that the largest absolute value is positive, and sort them lexicographically.
We then compute a cheap approximate interpolation, by interpolating the eigenvector matrices in Euclidean space and then down-projecting to a Stiefel manifold using a QR decomposition.
Graph interpolation is similarly difficult because, while sets have a true canonical ordering, graphs do not. 
To compute gradient penalty for graphs, we randomly rewire them with $p=0.1$, add Gaussian noise with the variance of $0.05$ to the resulting adjacency and clip it to a range of $[0,1]$. We do this for both, true and fake graphs.
On top of this, we found it helpful to also compute gradient penalty directly on the (unpermuted) true and fake samples of graphs and eigenvectors.

As QM9 has node and edge features, we add random perturbations to those categorical features. This is achieved by adding Gaussian noise and with certain probability randomly permuting the indices of the one-hot vector. We use the same variance and permutation probability as before.

Adding Gaussian noise to otherwise discrete values is important, as our generator uses either sigmoid or softmax to produce them and we pass the resulting output from the generator directly to the discriminator without hard sampling as we find that improves the stability of the training. 

\section{Spectral Graph Features (Wavelet)}
\label{app:spectral-mmd}

As graph generation is conditioned on the eigenvectors and eigenvalues, it is of interest to build a measure of spectral similarity between two different graphs. Therefore, we need the features to be a) invariant to node permutation, and b) descriptive for a specific range of eigenvalues.
Our construction takes inspiration from graph spectrograms~\citep[Chapter 3.2]{perraudin2017graph} and diffusion wavelet embeddings~\cite{donnat2018learning}. Given a collections of $P$ kernels $\phi_p: \mathbb{R}^+ \rightarrow \mathbb{R}$, we define spectral features for each node as: 
\begin{equation}
    \bm{S}[p, i] := \| [\phi(\bL)]_i \|_2^2 = \sum_{\ell=1}^n \phi(\lambda_\ell)^2 \bu_\ell^2[i] 
\end{equation}
We obtain invariance with respect of the node order by computing the histogram of the node independently of $p$
\begin{equation}
    \bar{\bm{S}}[p,q] = \text{hist}(\bm{S[p, \cdot]})[q]
\end{equation}
The Maximum Mean Discrepancy (MMD) can then be computed using the vectorized version of $\bar{\bm{S}}$.
We selected $P=12$ ab-spline wavelet functions using the PyGSP~\cite{pygsp} for the kernels $\phi_p$.

\section{Datasets}
\label{app:datasets}
As existing one-shot generative models only work with relatively small graphs~\cite{krawczuk2020gg, de2018molgan}, we take two commonly used datasets for their evaluation:

\noindent \textbf{Community-small Dataset \cite{you2018graphrnn}:} This synthetic dataset consists of $100$ random community graphs, which have between 12 and 20 nodes.

\noindent \textbf{QM9 \cite{ramakrishnan2014quantum}:} This dataset contains 134k organic molecules with up to 9 heavy atoms: carbon, oxygen, nitrogen, and fluorine. We follow the evaluation setup from~\cite{simonovsky2018graphvae, de2018molgan}, by using 10k molecules for validation, 10k molecules for testing, and the remaining ones for training.
Due to the very small size of graphs in this dataset, we reduce the number of PPGN layers used in the generator and discriminator to 3. As usually only validity, uniqueness and novelty are evaluated on this dataset. We use the number of valid, unique, and novel generated graphs for model selection instead of the MMD ratio.

To truly test our model's capability to generate large and valid graphs, we create two new challenging larger graph datasets:

\noindent \textbf{Planar Graph Dataset:} The dataset consists of $200$ planar graphs with $64$ nodes. The graphs are generated by applying Delaunay triangulation on a set of points that were placed uniformly at random.

\noindent \textbf{Stochastic Block Model Dataset:} We build $200$ Stochastic Block Model graphs, with $[2,5]$ communities (sampled at random) and $[20,40]$ nodes (sampled at random) in each of them. The inter-community edge probability is $0.3$ and the intra-community edge probability is $0.05$. To determine whether a generated graph is a valid SBM we employ the following procedure: We use Bayesian inference~\citep{peixoto2019bayesian}, to estimate communities present in the graph and then refine the community assignments using a merge-split Markov chain Monte Carlo scheme~\citep{peixoto2020merge}. From these assignments we recover the inter- and intra- community edge probabilities. We then use a Wald test~\cite{fahrmeir2007regression} to determine the probability that the predicted parameters match the original parameters. We only consider the graph valid, if the match probability is at least $0.9$. On top of that, we check if the generated graph has between 2 and 5 communities and if each of them has between 20 and 40 nodes. If the graph does not meet these restrictions we deem it invalid.

We also use a large real-world graph dataset:

\noindent \textbf{Protein Dataset~\citep{dobson2003distinguishing}:} This dataset consists of $918$ protein graphs, where each protein graph is constructed by connecting nodes (amino acids) by an edge if they are less than 6 Angstroms away. These protein graphs have between $100$ and $500$ nodes.

\section{Baselines}
\label{app:baselines}

For the GraphRNN~\citep{you2018graphrnn} and GRAN~\citep{liao2019efficient} baselines, we use the settings recommended by their authors. For GraphRNN we use the generation procedure proposed originally, where 1024 graphs are generated, and then the ones with the most similar node count to test graphs are selected. To be compatible with this, for GRAN and our models we directly condition on the test graph node distribution.

\section{Additional Background: Spectrum}
\label{app:spectrum}

\begin{figure}[t!]
\vskip 0.2in
\begin{center}
\centerline{\includegraphics[width=\linewidth]{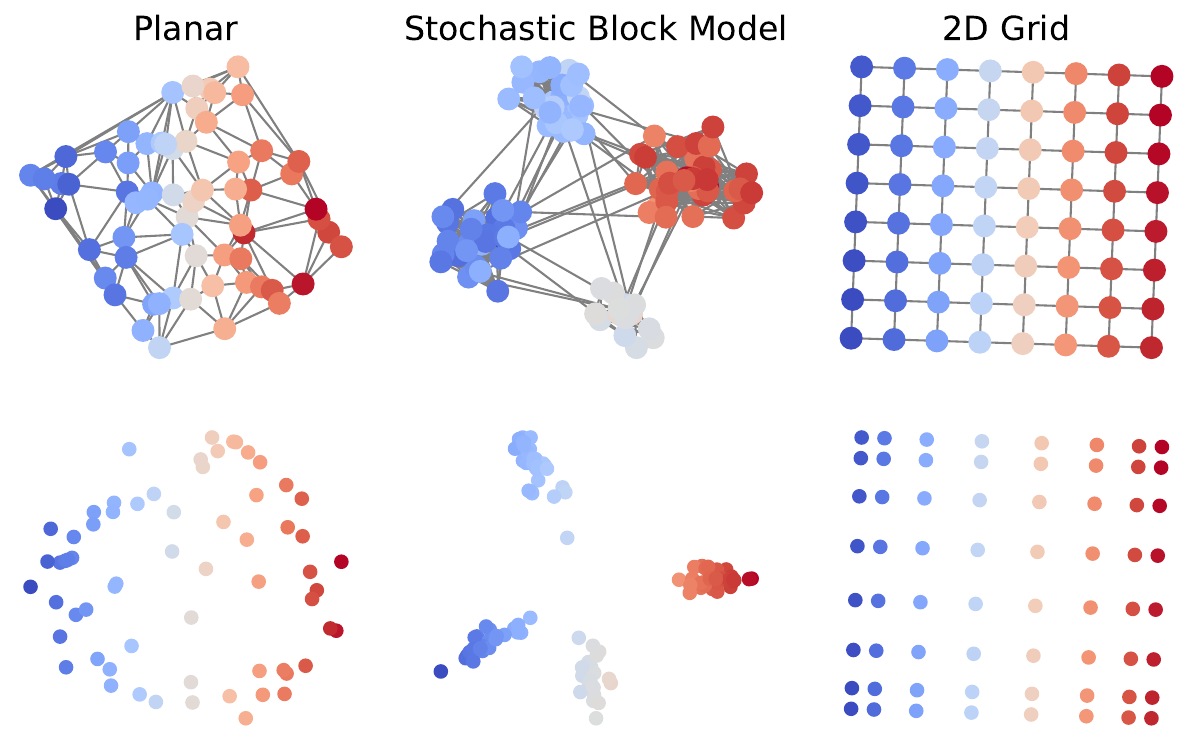}}
\caption{Normalized Laplacian eigenvectors which correspond to the lowest eigenvalues for different graphs. 
\textbf{Top}: Source graph. \textbf{Bottom}: The spectral embedding computed using the 2nd and 3rd eigenvectors of the graph normalized Laplacian captures the coarse graph structure. We do not use the 1st eigenvector as it only captures the degree distribution. Node color represents the value of the 2nd eigenvector.
}
\label{fig:intro-spectral}
\end{center}
\end{figure}

In the following, we provide further information about how the first few eigenvalues and eigenvectors convey and summarize properties relating to the global structure (Figure~\ref{fig:intro-spectral}): 

\textit{Connectivity.} The multiplicity of the zero eigenvalue equals the number of connected components~\cite{chung1997spectral}. 

\textit{Clusterability.} The differences between the smallest successive eigenvalues of connected graphs reveal how easy it is to partition the graph. The phenomenon is particularly prominent for the difference of the first two eigenvalues, often referred to as the spectral gap, which by the Cheeger inequality~\cite{alon1985lambda1, alon1986eigenvalues, SINCLAIR198993} upper and lower bounds the graph expansion and conductance. Nevertheless, analogous results also hold for higher-order eigenvalues~\cite{lee2014multiway}.

\textit{Diameter.} The first non-zero eigenvalue $\lambda_1$ can be used to bound the graph diameter as $\Delta \geq 1 / 2 m \lambda_1 $, with $m = |\cE|$ being the number of edges of $G$.

\textit{Node embeddings.} The first eigenvectors provide node embeddings that correlate with the geodesic distance.  
Eigenvectors have been used to embed the nodes of a graph (as in Laplacian Eigenmaps~\cite{belkin2003laplacian} and in Diffusion Maps~\cite{coifman2005geometric}) or to perform clustering (as in graph spectral clustering \cite{shi2000normalized}).

\textit{Link between the graph eigenvectors and the diffusion distance.} 
One way to link spectral embedding with graph nodes is to use the graph diffusion distance~\cite{hammond2013graph}. Given a decreasing kernel function $\phi(x)$ (such as $\phi(x) = e^{-t x }$ for example), one can define a diffusion distance between two vertices $v_i$ and $v_j$ as
\begin{align*}
    \mathcal{D}(v_{i},v_{j}) &:=   \left\|[\phi(\bL)]_{i,:} -[\phi(\bL)]_{j,:}\right\|_{2}^{2} \\ & =  \sum_{\ell=1}^n \phi\left(\lambda_{\ell}\right)^{2}\left(\bu_{\ell}[i]-\bu_{\ell}[j]\right)^{2}, 
\end{align*}
where $\phi(\bL) = \bU \phi(\bLambda) \bU^\top$.
Therefore, when embedding graph vertices using the eigenvectors associated with the $k$ lowest eigenvalues, one obtains a map where points are separated with a specific graph diffusion distance.

\section{Graph samples}
\label{appx:samples}
\begin{figure*}[t!]
\begin{center}

\centerline{\includegraphics[width=0.99\linewidth]{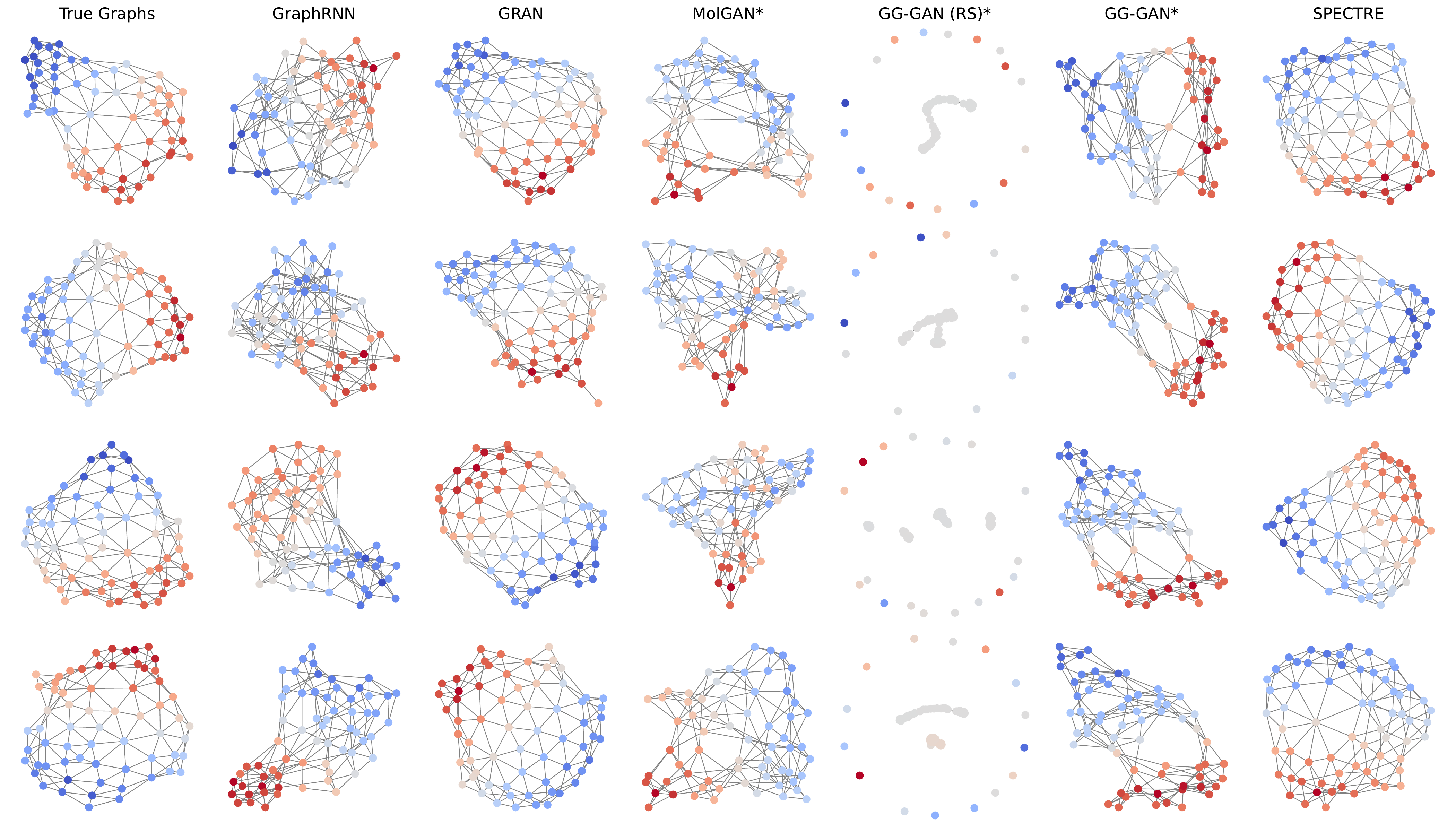}}
\caption{Uncurated set of sample Planar graphs produced by the models. Each row is conditioned on the same number of nodes.}
\label{fig:planar_samples}
\end{center}
\end{figure*}

\begin{figure*}[t!]
\begin{center}
\centerline{\includegraphics[width=0.99\linewidth]{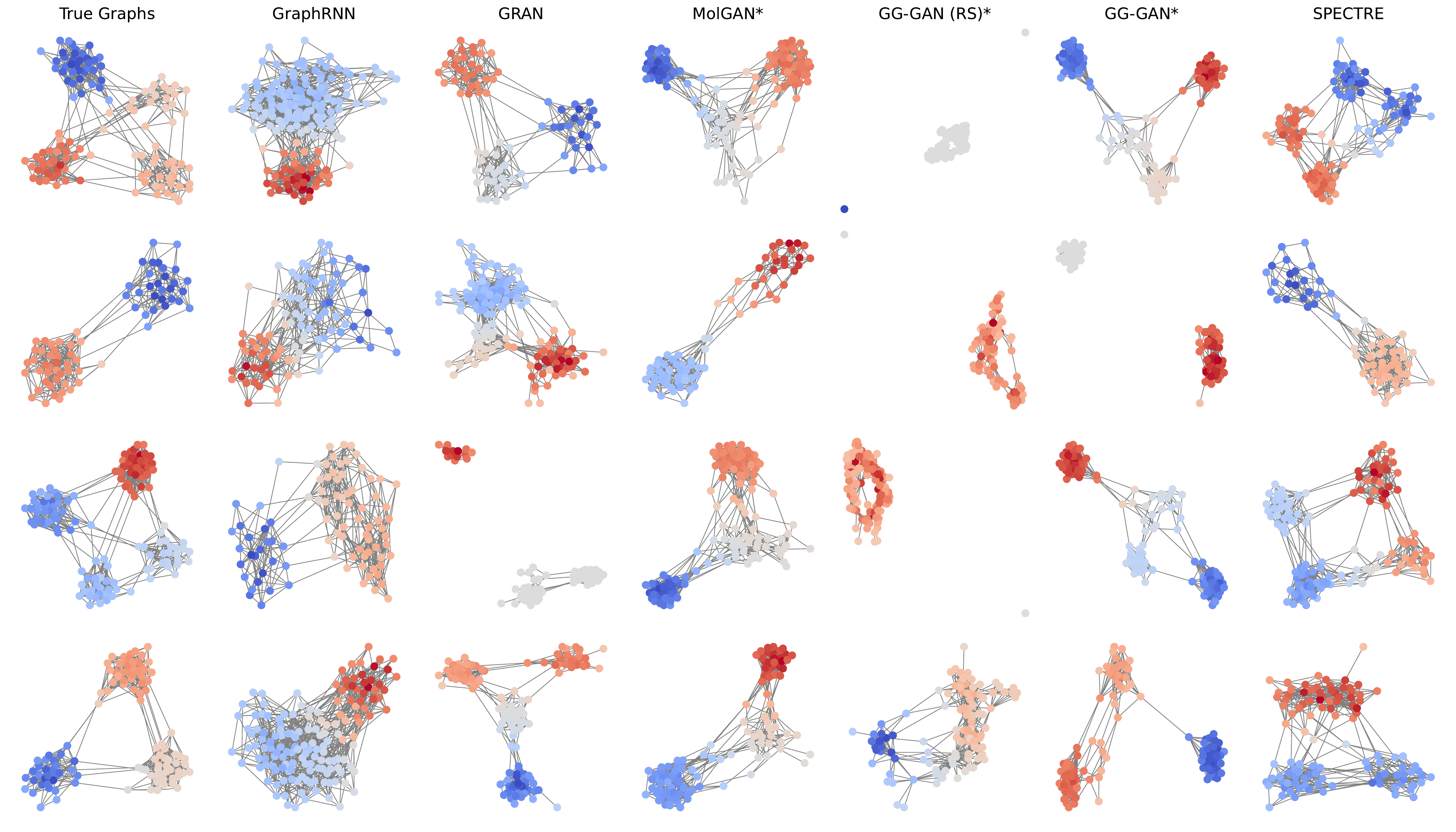}}
\caption{Uncurated set of sample Stochastic Block Model graphs produced by the models. Each row is conditioned on the same number of nodes.}
\label{fig:sb_samples}
\end{center}
\end{figure*}

\begin{figure*}[t!]
\begin{center}
\centerline{\includegraphics[width=0.99\linewidth]{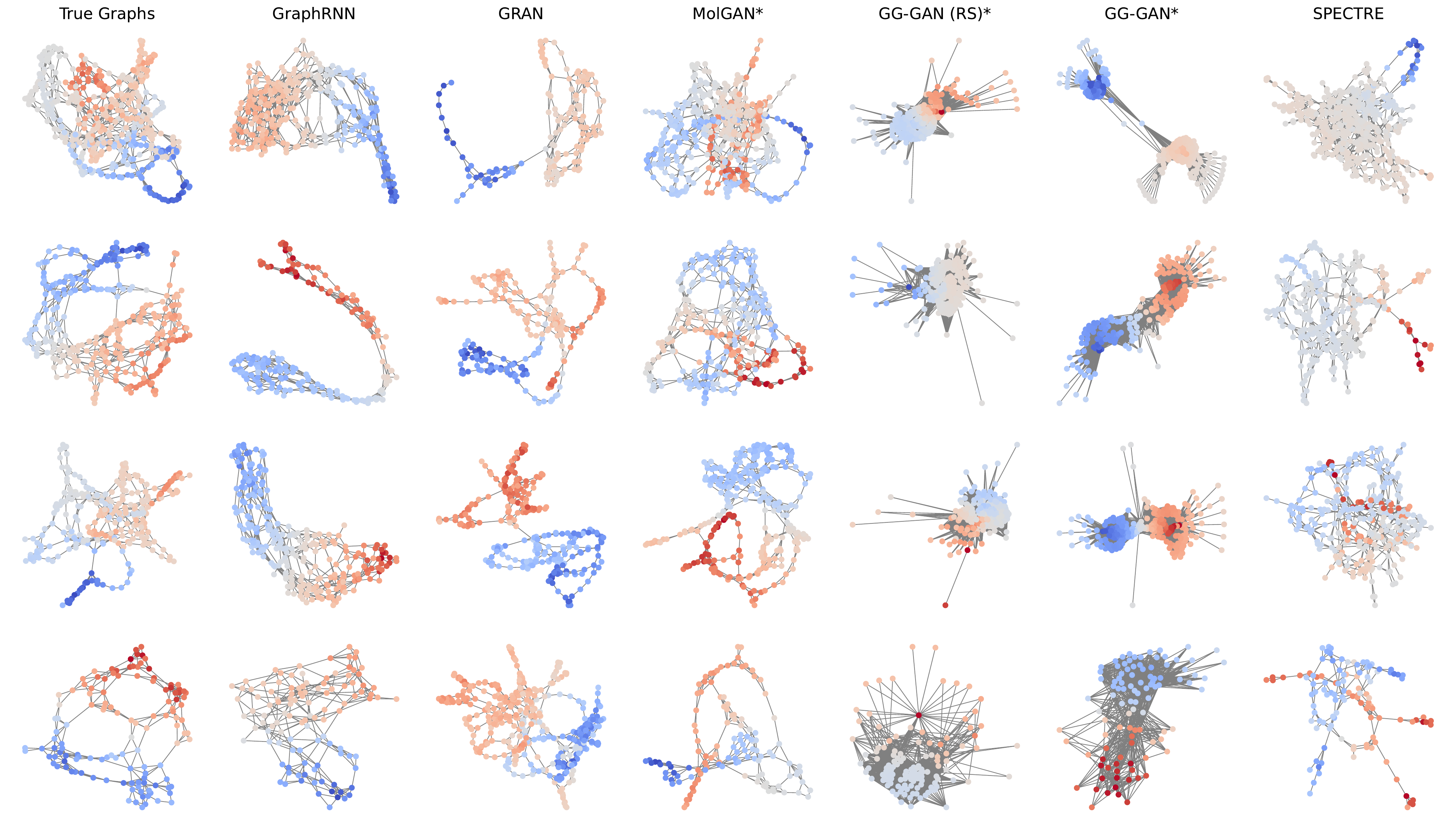}}
\caption{Uncurated set of sample Protein graphs produced by the models. Each row is conditioned on the same number of nodes.}
\label{fig:protein_samples}
\end{center}
\end{figure*}

\begin{figure*}[t!]
\begin{center}
\centerline{\includegraphics[width=0.99\linewidth]{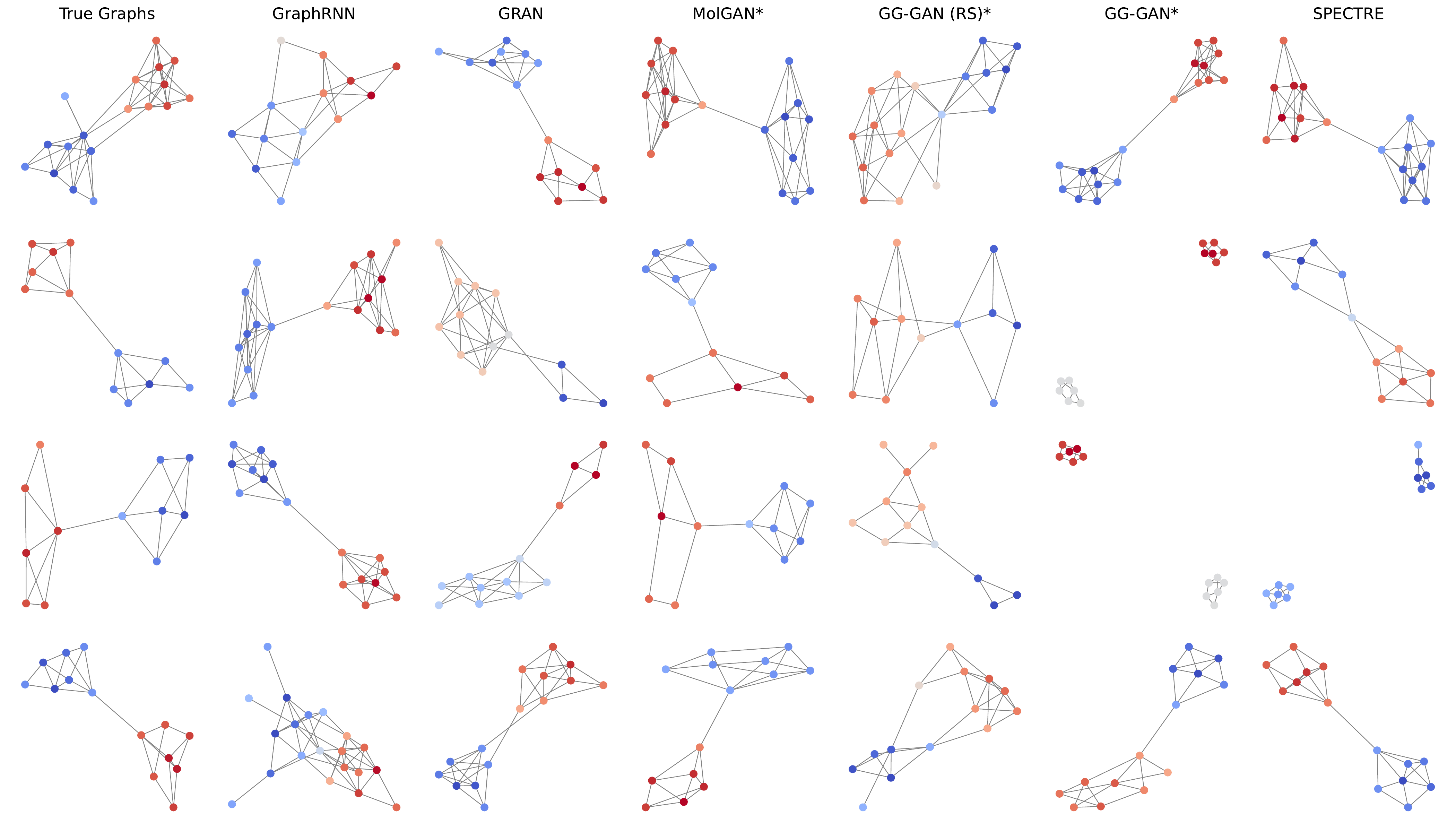}}
\caption{Uncurated set of sample Community graphs produced by the models. Each row is conditioned on the same number of nodes.}
\label{fig:community_samples}
\end{center}
\end{figure*}

\begin{figure*}[t!]
\begin{center}
\centerline{\includegraphics[width=0.94\linewidth]{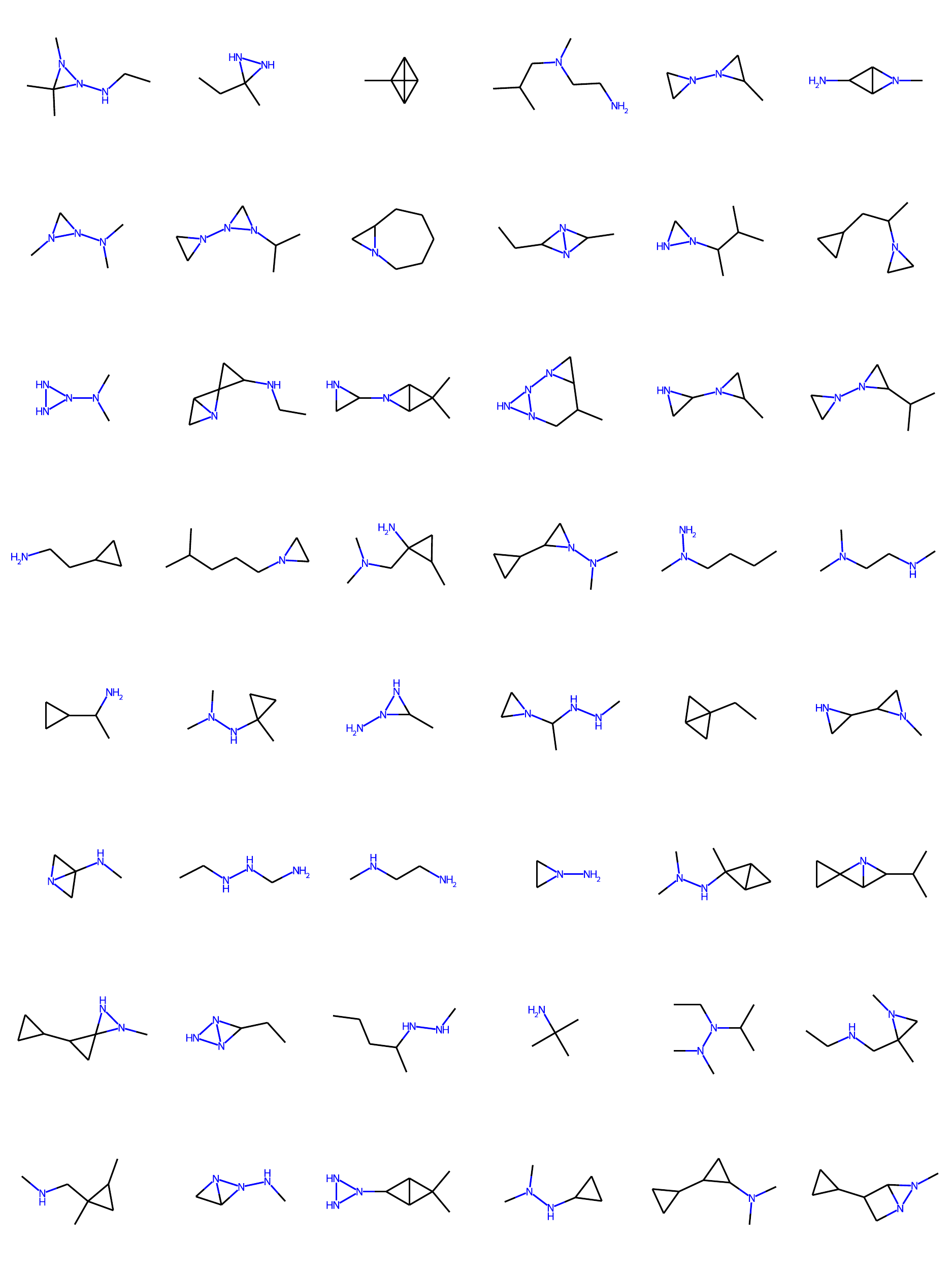}}
\caption{Uncurated set of unique sample QM9 molecules produced by SPECTRE.}
\label{fig:qm9_samples}
\end{center}
\end{figure*}

In this section, we showcase uncurated graph samples produced by the different models. Figure~\ref{fig:planar_samples}, \ref{fig:sb_samples}, \ref{fig:protein_samples} and \ref{fig:community_samples} displays respectively planar, SBM, protein and community graphs.  In every row, each model is conditioned on the same number of nodes. For protein graphs (Figure~\ref{fig:protein_samples}) all models fail to consistently produce connected graphs, thus following previous work \citep{you2018graphrnn,liao2019efficient} we display only the largest connected component. From these sample graphs, it is easy to see that spectral conditioning greatly improves the quality of the generated graphs and is the only model that truly respects hard constraints imposed on the training set, such as the maximum number of nodes allowed in an SBM component or graph planarity. We also provide uncurated unique molecule samples produced by SPECTRE in Figure~\ref{fig:qm9_samples}. While the given molecules are valid, in the sense that they can theoretically exist, they can be very unstable. It is also worth noting that while spectral conditioning prevents total mode collapse in this molecule generation task, the variety of the generated molecules is still somewhat low. Meaning, that while they are non-isomorphic they are mostly comprised of similar substructures.

\section{Adjacency matrices of MolGAN* proteins}
\label{appx:molgan_adj}

\begin{figure*}[t!]
\begin{center}
\centerline{\includegraphics[width=1.0\linewidth]{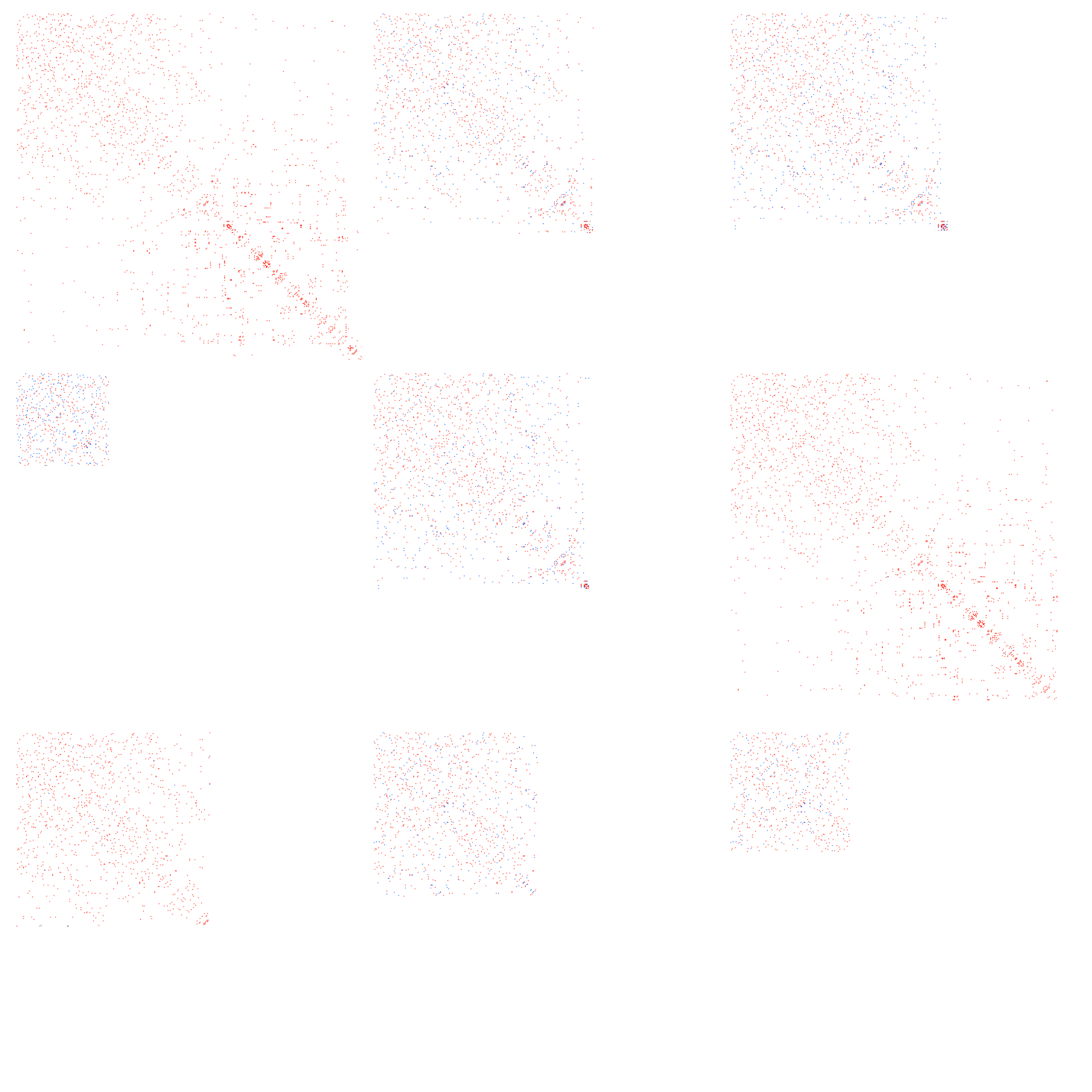}}
\caption{Adjacency matrices of 9 random proteins produced by MolGAN*. For each graph we highlight in red the edges that are also present in the first graph. All graphs are sub-graphs of the larger graph with a few changed edges.}
\label{fig:molgan_adjs}
\end{center}
\end{figure*}

In Figure~\ref{fig:molgan_adjs}, we show, that MolGAN* produces nearly identical adjacency matrices for different generated protein graphs. Observe, that while the node count differs in these adjacency matrices, as we cut the matrix based on the $n$ the model was conditioned on, resulting in overall non-isomorphic graphs, the edge connections are almost identical.

\begin{table}[h]
\centering
\resizebox{0.7\columnwidth}{!}{\begin{tabular}{@{}l*{2}{S}@{}}
\toprule
{Model} & {Mean Edit Distance}\\
\midrule  
{Test Dataset} & {78.8} \\
\midrule  
{GraphRNN}  & {83.7} \\
{GRAN}  & {48.8} \\
{MolGAN*}  & {17.6} \\
{GG-GAN (RS)*}  & {75.4}\\
{GG-GAN*}  & {90.9}\\
\rowcolor{Gray}
{\name ($k=16$)} & {97.0} \\
\bottomrule
\end{tabular}}
\caption{Mean edit distance as percentage of different edges between generated protein graphs. Consistent node ordering is assumed.}
\label{tab:edit_dist}
\end{table}  

To show this effect more clearly, we evaluate the percentage of different edges between any two graphs in the generated set in Table~\ref{tab:edit_dist}. The edit distance is computed assuming that the node ordering is consistent between the graphs. Given that all models, except GG-GAN (RS)*, either directly use an ordering or some (semi)-fixed initialization this is a reasonable assumption and the resulting value gives us the upper bound on the true mean edit distance. When comparing two graphs of different size, the larger graph is clipped to match the size of the smaller one. It is clear to see, that the mean edit distance between graphs generated by MolGAN* is very low and much lower than between graphs produced by any other model.

\end{document}